\documentclass{article}
\usepackage[preprint]{main}
\usepackage{amsmath,amsfonts,bm}
\usepackage{amsmath,amsfonts,bm}

\def\eqref#1{equation~\ref{#1}}

\def\1{\bm{1}}

\DeclareMathAlphabet{\mathsfit}{\encodingdefault}{\sfdefault}{m}{sl}
\SetMathAlphabet{\mathsfit}{bold}{\encodingdefault}{\sfdefault}{bx}{n}

\usepackage[table]{xcolor}
\usepackage[utf8]{inputenc}
\usepackage[T1]{fontenc}
\usepackage[colorlinks,
            linkcolor=red,
            anchorcolor=green,
            citecolor=blue
            ]{hyperref}
\usepackage{url}
\usepackage{textcomp}
\usepackage{booktabs}
\usepackage{amsfonts}
\usepackage{nicefrac}
\usepackage{microtype}
\usepackage{tabularx}
\usepackage{siunitx}
\usepackage{multirow}
\usepackage{booktabs}
\usepackage{capt-of}
\usepackage{microtype}
\usepackage{graphicx}
\usepackage{subfigure}
\usepackage{graphicx}
\usepackage{booktabs}
\usepackage{svg}
\usepackage{wrapfig}
\usepackage{caption}
\usepackage{algorithm}
\usepackage{algpseudocode}
\usepackage{listings}
\usepackage{makecell}
\usepackage{tcolorbox}
\tcbuselibrary{breakable}
\usepackage{makecell}
\usepackage{enumitem}

\usepackage{url}
\usepackage{flushend}
\usepackage{wrapfig}
\usepackage{placeins}
\usepackage{booktabs}
\usepackage{amsmath}
\usepackage{amssymb}
\usepackage{mathtools}
\usepackage{amsthm}
\usepackage{algorithm}
\usepackage{algpseudocode}
\definecolor{myred}{rgb}{0.8, 0.2, 0.2}
\definecolor{myblue}{rgb}{0.2, 0.2, 0.8}
\definecolor{TableTitleBg}{RGB}{63,63,63}
\definecolor{LightGrayBg}{RGB}{242,242,242}
\definecolor{LLMHeaderBg}{RGB}{220,220,220}
\definecolor{WhiteText}{RGB}{255,255,255}
\definecolor{BlackText}{RGB}{0,0,0}
\definecolor{RedText}{RGB}{255,0,0}

\usepackage[capitalize,noabbrev]{cleveref}
\definecolor{LeaderboardRowBlue}{RGB}{227,241,248}
\definecolor{LeaderboardGainGreen}{RGB}{0,150,80}
\definecolor{LeaderboardLossRed}{RGB}{190,45,65}
\colorlet{BOPDRowBlue}{LeaderboardRowBlue}
\colorlet{BOPDGainGreen}{LeaderboardGainGreen}
\colorlet{BOPDLossRed}{LeaderboardLossRed}

\makeatletter
\newcounter{scoreannot}
\newcommand{\scoreannotoverlays}{}
\newcommand{\scoreannot@add}[2]{
  \g@addto@macro\scoreannotoverlays{
    \node[
      anchor=mid west,
      inner sep=0pt,
      outer sep=0pt,
      xshift=0.10em,
      yshift=-0.10ex,
      font=\fontsize{4.9}{4.9}\selectfont
    ] at (#1.base east) {#2};
  }
}
\newcommand{\scoreannot}[2]{
  \stepcounter{scoreannot}
  \edef\scoreannotid{scoreannot\arabic{scoreannot}}
  \tikz[remember picture,baseline=(\scoreannotid.base)]{
    \node[inner sep=0pt, outer sep=0pt] (\scoreannotid) {#1};
  }
  \expandafter\scoreannot@add\expandafter{\scoreannotid}{#2}
}
\newcommand{\PrintScoreAnnotations}{
  \tikz[remember picture,overlay]{\scoreannotoverlays}
  \global\let\scoreannotoverlays\@empty
}
\makeatother

\title{Learning after Execution with Action-Specific Credit}
\author{
    Yufeng Tian$^{1}$,
    Jiahao Tang$^{1,2}$,
    Minghao Wu$^{1}$ \\
    $^1$The Chinese University of Hong Kong, Shenzhen, China\\
    $^2$Central South University, Changsha, China\\
    \texttt{tian\_yufeng@outlook.com}
}
\begin{document}
\pagestyle{plain}
\maketitle
\begin{abstract}
Execution is often used as terminal feedback: it scores a generated program, but the resulting state is not used for another learning action.
Learning after execution instead uses the execution result as an intermediate state that supports an additional learning interaction for the same policy.
This creates a credit-assignment problem because the two actions share the same final objective but play different roles in the trajectory.
Accordingly, action-specific credit assigns the initial action credit from the final trajectory outcome and the corrective action credit for improvement over the intermediate state.
Image-to-TikZ serves as an executable visual testbed for this setting.
On 945 held-out examples, learning after execution strengthens the shared policy under single-call inference and also produces a post-execution corrective capability.
Single-call render success reaches 92.70\%, compared with 88.25\% for continued single-stage GRPO after the same number of additional training iterations.
This corrective capability is not reproduced by continued single-stage training and depends on the intermediate state.
Alternative credit definitions also induce different group-relative preferences over the same trajectories.
Together, these results suggest that executable tasks contain useful learning structure beyond a terminal reward.
Code is available at
\href{https://github.com/tianyufengcn/Learning-after-Execution.git}{%
  \raisebox{-0.18\height}{\includegraphics[height=1em]{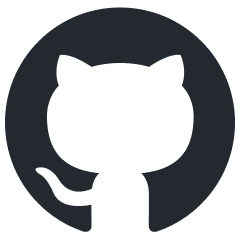}}%
  \,GitHub}.
\end{abstract}

\section{Introduction}
\label{sec:introduction}

Executing a generated program exposes the result of the model's prediction.
This result can be scored to train the policy, but it also provides a state on which the policy can act again~\citep{zheng2024opencodeinterpreter, li2025codeprm, bi2024iterative}.
In a single-stage interaction, execution marks the end of the trajectory.
The executed outcome provides a training signal, but the state produced by execution is not used for another learning action.
Rather than ending the interaction at this point, 
we ask whether the execution state can support an additional learning interaction, 
and whether learning from this interaction strengthens the shared policy itself.

This setting introduces a credit-assignment problem because the two actions share the same final objective but serve different roles in the trajectory~\citep{qu2026matchtir, du2026escaping}. 
The initial action generates the intermediate output and determines the state from which the corrective action begins, 
while the corrective action inherits that state and seeks to improve it. 
The final trajectory outcome is relevant to both actions, but the corrective action must also be evaluated relative to the state it receives. 
Training should therefore account for both where the trajectory ends and what changes after execution.

We study learning after execution by using the execution state of the policy's own prediction 
to create an additional learning interaction from the same training target. 
For each target, the shared policy first generates and executes an initial program, then revises or repairs it from the resulting state. 
We assign action-specific credit to the two actions.
The initial action receives credit from the final trajectory outcome, while the corrective action receives credit for improvement over the intermediate result. Both actions jointly update the same policy.

\begin{figure}[t]
    \centering
    \includegraphics[width=\linewidth]{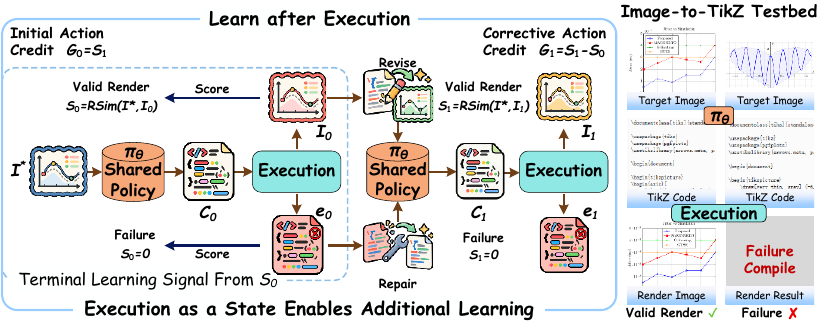}
    \caption{
        Overview of learning after execution with action-specific credit.
        Execution first provides a score for the initial action; learning after execution additionally uses the resulting render or failure state for a corrective action by the same policy.
        The initial action receives final-outcome credit, while the corrective action receives improvement-relative credit.
        The right panel illustrates the Image-to-TikZ instantiation with valid and failed executions.
        }
    \label{fig:overview}
\end{figure}

We use Image-to-TikZ as an executable visual testbed for this problem~\citep{belouadi2024detikzify,saito2025sketch2diagram}.
Image-to-TikZ reconstructs a target image as an executable TikZ program and belongs to the broader family of image-to-code~\citep{zhou2026omni,yang2025chartmimic,wu2025plot2code,si2025design2code,roberts2024image2struct}.
A generated program can be executed to reveal either a rendered image or an execution failure.
A valid render makes the visual difference from the target observable, while a failed execution identifies a program that still requires repair.
The result before and after correction can therefore be evaluated under the same execution and visual criteria.
This makes Image-to-TikZ suitable for studying both how a policy learns from an execution state and what changes when it acts on that state.

Using Image-to-TikZ, we study learning after execution through a series of controlled comparisons.
We separate improvement of the shared policy from the effect of invoking a corrective action at inference time, and distinguish learned post-execution correction from behavior that may arise from continued single-stage training alone.
We further examine how alternative credit definitions shape the group-relative learning signals assigned to the two actions.
Together, these analyses characterize whether learning after execution strengthens the shared policy, produces a distinct corrective capability, and how its learning is shaped by action-specific credit.
Our main contributions are:
\begin{itemize}
    \item
    We introduce learning after execution as a training setting in which execution is used not only to score a prediction, 
    but also to produce an intermediate state that provides additional learning experience for the same policy.
    \item
    We formulate action-specific credit for this setting, preserving a shared final-quality objective while matching credit to the different roles of initial generation and correction.

    \item
    We show that learning from post-execution interaction strengthens the shared policy even under single-call inference, 
    while also producing a corrective capability that is not reproduced by continued single-stage training alone.
\end{itemize}
\section{Preliminaries: Code Supervision and Execution Feedback}
\label{sec:code_to_execution}

Before introducing learning after execution, we define the program generation,
execution score, and single-stage training setting used throughout the paper.
In our Image-to-TikZ instantiation, a target image $I^\ast$ is mapped to an
executable TikZ program $C$.
Executing $C$ either produces a rendered image or results in execution failure.
These outcomes define the execution score used by Direct-GRPO and by our method
in Section~\ref{sec:workflow}.

\subsection{Supervised Image-to-TikZ}
\label{sec:sft}

Given paired training data $\mathcal{D}=\{(I_i^\ast,C_i^\ast)\}_{i=1}^{N}$,
where $I_i^\ast$ is a target image and $C_i^\ast$ is its reference TikZ program,
supervised fine-tuning trains the model to generate $C$ conditioned on $I_i^\ast$.
The supervision is defined on the reference program rather than on the executed
outcome of a program sampled from the policy.
Different TikZ programs can produce similar rendered figures, while small code
changes can alter the visual result or break execution.
The rendered consequence of the model's own prediction therefore does not
directly participate in the supervised training objective.

\subsection{Direct Reinforcement Learning with Execution Feedback}
\label{sec:direct_grpo}

Execution exposes the consequence of a generated program.
For each target image, we sample a group of $K$ programs
$C_i \sim \pi_{\theta_{\mathrm{old}}}(\cdot \mid I^\ast)$,
for $i=1,\ldots,K$, and evaluate their executed outputs.
We define the execution score

\begin{equation}
S(C;I^\ast)
=
\begin{cases}
\operatorname{RSim}\!\left(\mathcal{E}(C),I^\ast\right),
&
\text{if } C \text{ produces a valid render},
\\[2pt]
0,
&
\text{otherwise}.
\end{cases}
\label{eq:execution_score}
\end{equation}

$\mathcal{E}$ denotes the \LaTeX{} execution and rasterization pipeline.
Following TikZilla, we use RSim as the visual similarity reward,
which compares patch-level representations from the DeTikZify-V2 image encoder
using Earth Mover's Distance~\citep{kusner2015word,belouadi2024detikzify,greisinger2026tikzilla}.
The score combines executability and visual fidelity:
failed executions receive zero, while valid renders are scored by their
similarity to the target.

For Direct-GRPO, each sampled program receives
$S_i=S(C_i;I^\ast)$.
Within each target-image group, we mean-center these scores to obtain
group-relative advantages and optimize them with the clipped GRPO objective~\citep{shao2024deepseekmath}.
Implementation details are given in Appendix~\ref{app:workflow_details}.
We refer to this single-stage setting as Direct-GRPO.

Direct-GRPO learns from the executed outcomes of programs sampled from the
current policy.
Its optimization signal reflects both executability and rendered visual
fidelity.
Execution, however, still marks the end of the interaction:
the result is used to score the sampled program, but the policy does not observe
the rendered output or failure state as input for another action.
The next section uses this execution result as an intermediate state on which
the same policy can act again.
\section{Learning after Execution with Action-Specific Credit}
\label{sec:workflow_grpo}

Direct-GRPO uses execution as terminal feedback.
We instead use the execution result as an intermediate state on which the same
policy can act again.
The policy first generates and executes a program, then revises or repairs it
based on the observed execution result.
The initial and corrective actions are jointly trained with action-specific
credit.

\subsection{Execution-Conditioned Correction}
\label{sec:workflow}

Let $I^\star$ denote the target image.
At training iteration $k$, the same policy $\pi_{\theta_k}$ is
called twice within each trajectory.
The first call directly generates an initial TikZ program,

\begin{equation}
C_{0,i}
\sim
\pi_{\theta_k}
\left(
\cdot \mid I^\star
\right),
\label{eq:stage0_policy}
\end{equation}

where $i$ indexes one sampled trajectory.
We execute $C_{0,i}$ using the same \LaTeX{} rendering pipeline introduced in
Section~\ref{sec:direct_grpo} and compute

\begin{equation}
S_{t,i}
=
S(C_{t,i};I^\star),
\qquad
t\in\{0,1\},
\label{eq:workflow_score}
\end{equation}

using the execution score defined in Section~\ref{sec:direct_grpo}.
Thus, $S_{0,i}$ evaluates the intermediate program produced by the first call.
The score is used for credit computation; it is not itself the observation
provided to the second call.

The execution result determines the context for the corrective action.
If $C_{0,i}$ produces a valid rendered image $I_{0,i}$,
the model receives the target image, the current render, and the current TikZ
program.
It can therefore compare the target with what its own program actually
produced.
If no valid render is produced, the model instead receives the target image,
the current TikZ attempt, and a compact execution status $e_{0,i}$.
The corrective action is sampled as

\begin{equation}
C_{1,i}
\sim
\begin{cases}
\pi_{\theta_k}
\left(
\cdot \mid I^\star,I_{0,i},C_{0,i}
\right),
&
\text{valid Stage~0 render},
\\[3pt]
\pi_{\theta_k}
\left(
\cdot \mid I^\star,C_{0,i},e_{0,i}
\right),
&
\text{otherwise}.
\end{cases}
\label{eq:stage1_policy}
\end{equation}

The first branch performs visual revision.
The second branch repairs, completes, or replaces an unsuccessful program.
We then execute $C_{1,i}$ and obtain the final score $S_{1,i}$ using
Eq.~\ref{eq:workflow_score}.

Stage~0 and Stage~1 are two calls to the same policy $\pi_{\theta_k}$ under
different contexts.
The first call observes only the target image.
The second call observes the execution state produced by the first action.
No parameter update occurs between the two calls.
Both actions are sampled before the shared policy is updated.

\subsection{Action-Specific Credit from Incremental Rewards}
\label{sec:workflow_credit}

Both actions serve the same final-quality objective,
$J(\theta)=\mathbb{E}_{\tau\sim\pi_\theta}[S_1]$.
With two actions and no discounting, we decompose the execution score into
incremental rewards and use the corresponding returns-to-go as
action-specific credit:

\begin{equation}
\begin{aligned}
r_{0,i} &= S_{0,i},
&\qquad
r_{1,i} &= S_{1,i}-S_{0,i},
\\
G_{0,i} &= r_{0,i}+r_{1,i}=S_{1,i},
&\qquad
G_{1,i} &= r_{1,i}=S_{1,i}-S_{0,i}.
\end{aligned}
\label{eq:workflow_returns}
\end{equation}

The incremental rewards telescope to the final score,
$r_{0,i}+r_{1,i}=S_{1,i}$.
The initial action therefore receives the return of the complete trajectory,
while the corrective action receives the improvement relative to the
intermediate state it inherits.

For the initial action, assigning only $S_{1,i}-S_{0,i}$ would omit its
immediate reward $S_{0,i}$ and, for a fixed final score, favor a lower
intermediate score.
We therefore use the complete return $G_{0,i}=S_{1,i}$ and evaluate the
initial action by the final quality reached by the trajectory.

For the corrective action,
$G_{1,i}=S_{1,i}-S_{0,i}$ directly measures the score change from the state it
receives.
Its raw credit is positive when the execution score increases and negative
when the score decreases.
Because unsuccessful executions receive zero,
the same definition also covers execution repair and failure:
if $S_{0,i}=0$, then $G_{1,i}=S_{1,i}\geq0$;
if $S_{1,i}=0$, then $G_{1,i}=-S_{0,i}\leq0$.
Visual revision and execution repair are therefore handled by the same
credit definition.

\subsection{Group-Relative Optimization}
\label{sec:workflow_optimization}

For each target image, we sample $K$ complete trajectories, with $K=8$ in all
experiments using the corrective action.
Each trajectory
$\tau_i=(C_{0,i},S_{0,i},C_{1,i},S_{1,i})$
contains an independently sampled initial action followed by a corrective
action conditioned on its own execution state.
Thus, a group contains $K$ independently sampled trajectories rather than one
shared Stage~0 program followed by multiple Stage~1 candidates.

The action-specific returns in Eq.~\ref{eq:workflow_returns} are centered
separately within the same target-image group.
With the implementation scale $\lambda=10$,

\begin{equation}
A_{t,i}
=
\lambda
\left(
G_{t,i}
-
\frac{1}{K}\sum_{j=1}^{K}G_{t,j}
\right),
\qquad
t\in\{0,1\}.
\label{eq:workflow_advantages}
\end{equation}

We subtract only the group mean and do not normalize by the group standard
deviation.
$A_{0,i}$ is assigned to the completion tokens of $C_{0,i}$, while
$A_{1,i}$ is assigned to those of $C_{1,i}$.

The distinction between raw credit and group-relative advantage is important.
A positive $G_{1,i}$ means that the corrective action increases the execution
score, but it does not necessarily imply $A_{1,i}>0$;
the latter also depends on the mean improvement within the sampled group.

For the corrective action, $S_{0,i}$ is determined before $C_{1,i}$ is
sampled and acts as a state-dependent baseline.
Appendix~\ref{app:credit_analysis} further compares improvement-relative and absolute final-score
credit.

The group-relative objective is applied separately to the two action sets,

\begin{equation}
\mathcal{L}
=
0.5\mathcal{L}_{\mathrm{GRPO}}(C_0,A_0)
+
0.5\mathcal{L}_{\mathrm{GRPO}}(C_1,A_1).
\label{eq:workflow_loss}
\end{equation}

Both terms optimize the same policy parameters.
All Stage~0 and Stage~1 actions in a training iteration are sampled from the
same policy $\pi_{\theta_k}$ before any parameter update.
The corresponding losses for the two stages are optimized sequentially over action microbatches using the shared policy.
\section{Experiments}
\label{sec:experiments}

The experiments examine how using execution during training affects both
initial generation and subsequent correction.
We first evaluate overall performance under direct and post-execution inference.
We then study the corrective action, the role of action-specific credit,
and how correction varies across executed intermediate states.

\subsection{Experimental Setup}
\label{sec:experimental_setup}

\paragraph{Models and Training Data.}
Qwen3-VL-8B-Instruct~\citep{bai2025qwen3} is used as the base model throughout the experiments.
All supervised and reinforcement learning data are drawn from DaTikZ-V4~\citep{greisinger2026tikzilla}.
Direct-SFT is trained on approximately 4K Image-to-TikZ pairs.
Direct-GRPO is initialized from the final SFT checkpoint and trained on 1K samples for 125 training iterations.
Our method is initialized from the final Direct-GRPO checkpoint and trained on the same 1K samples for 125 training iterations.

\paragraph{Evaluation Data.}
A temporally held-out evaluation set is constructed using the same source collection procedure as DaTikZ-V4, restricted to data available from 2026 onward.
Samples are collected from arXiv, GitHub, and TeX StackExchange.
The 1,095 examples are divided into two non-overlapping subsets.
\textbf{Dev-150} contains 150 examples for checkpoint selection, ablations, and behavioral analysis.
\textbf{Final-945} contains the 945 examples and is reserved for final evaluation.

\paragraph{Evaluation Metrics.}
Render rate measures whether the generated TikZ program executes successfully.
Visual fidelity is evaluated with RSim~\citep{greisinger2026tikzilla}, DINO~\citep{caron2021emerging}, DreamSim~\citep{fu2023DreamSim}, and CLIPImg~\citep{radford2021learning}.
RSim is also used as the reinforcement learning reward.
Unless stated otherwise, visual metrics use the \emph{fixed-zero} convention, where failed executions receive a score of zero.

\paragraph{Inference Protocol.}
All models are evaluated with temperature $1.0$, top-$p$ $0.9$, repetition penalty $1.0$, and at most 8192 generated tokens.
Generation stops at the first complete \verb|\end{document}|.
Stage~0 denotes the initial output produced from the target image alone,
while Stage~1 denotes the raw output produced after observing the Stage~0 execution state.
We additionally evaluate an execution-aware selection rule over the two outputs already produced.
If one stage fails to render, the valid output is retained.
If both render, the output with the higher RSim score is selected, with ties assigned to Stage~0.
No additional model generation is required by the selection rule.

\subsection{Overall Results}
\label{sec:main_results}

We first compare the main models on Final-945.
Our initializer is the final Direct-GRPO checkpoint, evaluated with the same
Stage~0/Stage~1 interaction before further training.
Because decoding is stochastic, its Stage~0 result and the separately sampled
Direct-GRPO baseline can differ slightly.
As an additional control, we continue Direct-GRPO from the same checkpoint
on the same training set for another 100 training iterations.

\begin{table*}[th]
    \caption{
        Main results on Final-945.
        Visual metrics use the fixed-zero convention.
        \emph{Direct} denotes single-call generation.
        \emph{Stage~0} is the initial single-call output of our policy,
        while \emph{Stage~1} is produced after execution-conditioned correction.
        \emph{Selected} chooses between the two outputs using execution success and RSim.
        Direct-GRPO (+100) continues single-stage GRPO for 100 additional training iterations
        from the same Direct-GRPO checkpoint.
        Best results are shown in bold and second-best results are underlined.
    }
    \label{tab:final945}
    \centering
    \setlength{\tabcolsep}{5.5pt}
    \begin{tabular}{llccccc}
        \toprule
        \makecell[l]{Method}
        & \makecell[l]{Output}
        & \makecell[c]{Render\\Success}  $\uparrow$
        & \makecell[c]{RSim $\uparrow$}
        & \makecell[c]{DINO $\uparrow$}
        & \makecell[c]{DreamSim $\uparrow$}
        & \makecell[c]{CLIPImg $\uparrow$} \\
        \midrule
        
        Base model
        & Direct
        & 71.64\%
        & 0.4033
        & 0.6025
        & 0.5545
        & 0.6435 \\

        SFT
        & Direct
        & 64.66\%
        & 0.3554
        & 0.5427
        & 0.4960
        & 0.5781 \\

        Direct-GRPO
        & Direct
        & 81.90\%
        & 0.4709
        & 0.7046
        & 0.6468
        & 0.7410 \\

        Direct-GRPO (+100)
        & Direct
        & 88.25\%
        & 0.5135
        & 0.7629
        & 0.7019
        & 0.7996 \\

        \midrule

        Ours (initializer)
        & Stage 0
        & 83.07\%
        & 0.4766
        & 0.7129
        & 0.6544
        & 0.7489 \\

        Ours (initializer)
        & Stage 1
        & 83.17\%
        & 0.4781
        & 0.7142
        & 0.6556
        & 0.7501 \\

        Ours
        & Stage 0
        & 92.70\%
        & 0.5382
        & 0.8014
        & 0.7330
        & 0.8372 \\

        Ours
        & Stage 1
        & \underline{92.80\%}
        & \underline{0.5446}
        & \underline{0.8046}
        & \underline{0.7401}
        & \underline{0.8409} \\

        Ours
        & Selected
        & \textbf{94.18\%}
        & \textbf{0.5649}
        & \textbf{0.8221}
        & \textbf{0.7593}
        & \textbf{0.8583} \\

        \bottomrule
    \end{tabular}
\end{table*}

Table~\ref{tab:final945} shows that SFT does not improve performance on this held-out set,
whereas Direct-GRPO reaches a render rate of $81.90\%$,
together with consistent improvements across all four visual metrics.
Optimizing executed outcomes provides a substantially stronger signal than reference code supervision alone.

Continuing Direct-GRPO for another 100 training iterations provides further gains,
raising render success to $88.25\%$.
Starting from the same Direct-GRPO checkpoint and after the same number of additional training iterations, our Stage~0 output reaches $92.70\%$
and is higher under all four visual metrics.
Continued single-stage training therefore does not reproduce the gain obtained with our training.
Importantly, this gain is already present in Stage~0, which uses the same
single-call generation setting as the direct baselines.

Simply adding a second execution-conditioned call is insufficient.
At initialization, Stage~0 and Stage~1 differ little.
After training, raw Stage~1 reaches $92.80\%$ render success and achieves slightly higher visual similarity than Stage~0 across all four metrics.
Execution-aware selection further raises render success to $94.18\%$ and improves all four visual metrics.

These results separate two effects.
Training strengthens the shared policy itself, as shown by the improved Stage~0 output,
while the second action provides an additional opportunity to act on the execution state.
We examine the latter behavior next.

\subsection{Learning Dynamics of Generation and Correction}
\label{sec:workflow_correction}

Since Stage~0 already improves substantially, we isolate the corrective action itself.
For visual correction, we consider only examples where both stages render successfully,
termed the \emph{paired success} subset.
For a visual similarity metric $M$, we define

\begin{equation}
    \Delta M
    =
    M(I_1,I^\star)-M(I_0,I^\star),
    \label{eq:strict_delta}
\end{equation}

where $I_0$ and $I_1$ are the rendered outputs before and after correction, and $I^\star$ is the target image.
This comparison excludes gains from execution recovery and isolates the visual change produced by Stage~1 on already executable outputs.
For the correction-outcome composition, we use a small neutral region:
$\Delta\mathrm{RSim}>0.005$ is counted as improvement,
$|\Delta\mathrm{RSim}|\leq0.005$ as neutral,
and $\Delta\mathrm{RSim}<-0.005$ as degradation.

\begin{figure}[ht]
    \centering
    \includegraphics[width=\linewidth]{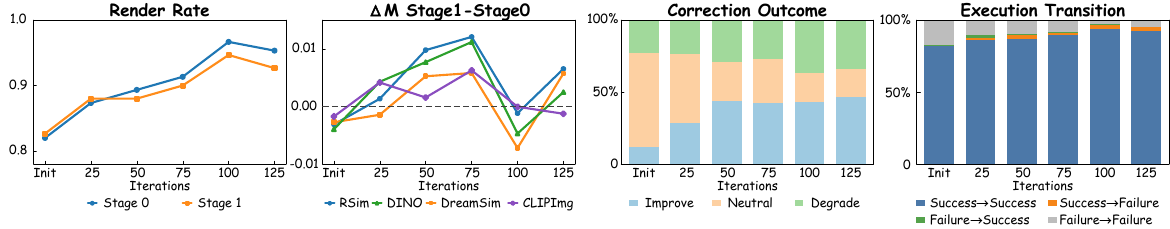}
    \caption{
        Learning dynamics of our method on Dev-150.
        The figure tracks Stage~0 and Stage~1 render rates,
        paired visual changes $\Delta M$,
        the distribution of correction outcomes,
        and execution transitions across training checkpoints.
    }
    \label{fig:workflow_dynamics}
\end{figure}

Figure~\ref{fig:workflow_dynamics} shows that the initializer does not yet act as a reliable visual corrector.
On paired-success examples, its average Stage~1 change is negative under all four visual metrics.
Strong direct generation therefore does not automatically provide the ability to revise an executed result.
Further single-stage optimization shows the same pattern:
when Continued Direct-GRPO is evaluated with the same execution-conditioned second call,
its paired visual changes remain close to zero.

This behavior changes during training.
At step 75, the average Stage~1 change is positive under RSim, DINO, DreamSim, and CLIPImg.
RSim is used for training.
The concurrent gains under the other three metrics show that the visual improvement is not confined to the reward metric.
Correction quality is not monotonic across training, even though render rate improves much more consistently.
This shows that initial generation quality and the effect of the corrective action can evolve differently.
We return to this difference in Section~\ref{sec:state_dependent}.

The correction-outcome distribution also changes substantially.
The fraction of paired-success examples improved by Stage~1 rises from $12.2\%$ for the initializer
to $42.5\%$ at step 75 and $46.8\%$ at step 125.
Most successful Stage~0 executions remain executable after Stage~1,
while failed-to-successful recovery remains relatively rare on Dev-150.
Together, these results show that training changes both the initial action
and the behavior of the subsequent corrective action.
Checkpoint 100 is selected on Dev-150 for subsequent behavioral analyses and final evaluation on Final-945,
while training dynamics are examined across the full checkpoint trajectory.

\subsection{Credit Assignment in Post-Execution Learning}
\label{sec:credit_ablation}

\begin{wraptable}[10]{r}{0.48\linewidth}
    \vspace{-\intextsep}
    \caption{
        Factorized ablation of action-specific credit assignment.
        Rows vary Stage~0 credit and columns vary Stage~1 credit.
    }
    \label{tab:credit_design}
    \centering
    \setlength{\tabcolsep}{5pt}
    \begin{tabular}{lcc}
        \toprule
        & \multicolumn{2}{c}{Stage~1 credit} \\
        \cmidrule(lr){2-3}
        Stage~0 credit
        & $\lambda(S_1-S_0)$
        & $\lambda S_1$ \\
        \midrule

        $\lambda S_1$
        & \textbf{A (Ours)}
        & C \\

        $\lambda S_0$
        & B
        & D \\

        \bottomrule
    \end{tabular}
\end{wraptable}

We factorize the action-specific credit design into a $2\times2$ ablation.
Stage~0 credit is either the final trajectory score $S_1$ or the local intermediate score $S_0$,
while Stage~1 credit is either the improvement $S_1-S_0$ or the absolute final score $S_1$.
This yields the four variants in Table~\ref{tab:credit_design}, with A corresponding to our method.

\paragraph{Downstream versus local credit for the initial action.}

The A--B comparison isolates Stage~0 credit.
At step 50, B already achieves a slightly higher Stage~0 render rate than A,
yet its subsequent Stage~1 correction is nearly neutral.
A produces a substantially larger correction at the same checkpoint
and remains positive at step 75, whereas B becomes slightly negative.
Thus, optimizing the intermediate output itself does not produce the same subsequent correction behavior.
Using downstream credit $S_1$ instead favors initial actions according to the quality reached after the corrective action.

\paragraph{Improvement-relative versus absolute credit for the corrective action.}

The A--C comparison isolates Stage~1 credit.
We recompute absolute $S_1$-based Stage~1 advantages on the 8,000 trajectories collected from the completed A run.
The difference is substantial at both trajectory and group levels:
replacing $S_1-S_0$ with $S_1$ flips the sign of the group-relative advantage for $37.2\%$ of trajectories
and changes the highest-advantage trajectory in $74.5\%$ of target-image groups.
The two credit definitions therefore induce substantially different group-relative preferences over corrective actions.

\begin{figure}[ht]
    \centering
    \includegraphics[width=\linewidth]{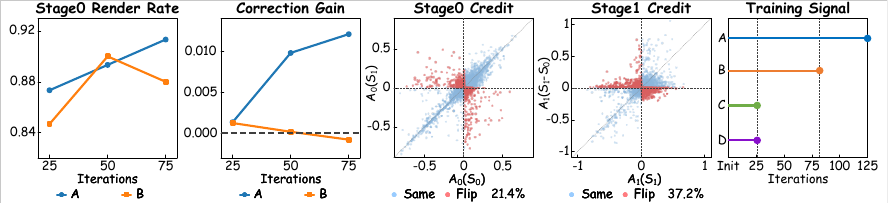}
    \caption{
        Effect of action-specific credit assignment.
        The figure compares Stage~0 quality and correction gain under alternative credit definitions,
        measures how the resulting group-relative preferences differ,
        and tracks the training signal of all four ablation variants.
    }
    \label{fig:credit_ablation}
\end{figure}

\paragraph{Optimization behavior of the credit variants.}

Variants C and D share absolute Stage~1 credit and enter the same failure regime from step 26
despite using different Stage~0 credits.
Their Stage~1 generations become substantially longer and normal document termination deteriorates,
causing most outputs to fail execution.
For D, the average Stage~1 completion approaches 7K tokens and $86\%$ of completions reach the generation limit.

Once both stage scores collapse to zero within a group,
reward variation disappears and the group-relative advantages,
policy loss, and gradient norm become zero.
The zero learning signal therefore follows the generation failure rather than preceding it.
In contrast, A and B remain executable and retain nonzero training signals on the same groups.
C and D differ in Stage~0 credit but share the same absolute Stage~1 credit,
making this the common factor associated with their early failure.
Together, the fixed-trajectory preference analysis and observed optimization behavior
show that $S_1$ and $S_1-S_0$ are not interchangeable credit signals for the corrective action.

\subsection{Correction Outcomes Across Intermediate States}
\label{sec:state_dependent}

\begin{figure}[ht]
    \centering
    \includegraphics[width=\linewidth]{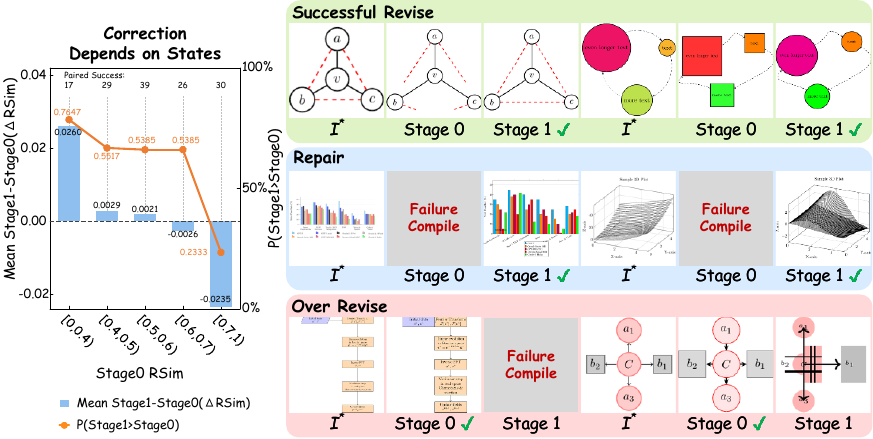}
    \caption{
        Correction behavior as a function of Stage~0 quality.
        We report the mean Stage~1 change in RSim and the frequency with which
        Stage~1 achieves higher RSim than Stage~0 across Stage~0 quality ranges,
        together with qualitative examples of beneficial and harmful corrections.
    }
    \label{fig:state_dependent}
\end{figure}

The previous analysis shows that the effect of the corrective action is not constant across training.
We therefore examine how Stage~1 outcomes vary with the quality of the executed Stage~0 state.
Using the Dev-selected checkpoint at step 100,
Figure~\ref{fig:state_dependent} shows a clear pattern across intermediate-state quality.
Stage~1 provides the largest average gain when the intermediate render is weak.
For Stage~0 RSim below $0.4$, it improves over Stage~0 in $76\%$ of paired examples.
As Stage~0 quality increases, both the average correction gain
and the probability of improvement generally decrease.
In the strongest group, the average change becomes negative
and Stage~1 improves over Stage~0 in only $23\%$ of cases.
Additional qualitative examples covering successful revision, execution repair, over-revision, preservation, 
and unresolved cases are provided in Appendix~\ref{app:qualitative}.

This pattern helps explain the non-monotonic correction curves in Figure~\ref{fig:workflow_dynamics}.
Weak intermediate renders leave larger discrepancies for Stage~1 to address.
As Stage~0 becomes stronger, the remaining errors become smaller
and further modification carries a greater risk of degrading an already good reconstruction.
Thus, a stronger Stage~0 policy can reduce the average marginal value of correction
even when Stage~1 remains capable of improving weak states.

\paragraph{Execution-Aware Selection.}

The asymmetric effect of correction motivates selecting between the two outputs already produced by the policy.
We use the execution-aware rule defined in Section~\ref{sec:experimental_setup},
which requires no additional model generation.
On Dev-150, checkpoint 100 gives the strongest selected result.
Selection raises render success to $97.3\%$,
compared with $96.7\%$ for Stage~0 and $94.7\%$ for raw Stage~1,
while RSim-FZ reaches $0.5646$.
DINO, DreamSim, and CLIPImg also improve although they are not used by the selector.

The checkpoint and selection rule are then frozen before evaluation on Final-945.
As shown in Table~\ref{tab:final945},
selection raises render success from $92.80\%$ for raw Stage~1 to $94.18\%$
and improves all four fixed-zero visual metrics.
Stage~0 can therefore be used directly as the single-call output,
while the post-execution action and selection provide an optional additional refinement mode.
\section{Related Work}
\label{sec:related_work}

\paragraph{Rendering- and execution-aware reinforcement learning.}
RLRF~\citep{rodriguez2026rendering} optimizes vector graphics with rewards from rendered SVG outputs, while TikZilla~\citep{greisinger2026tikzilla} applies GRPO to Text-to-TikZ using rendered visual similarity.
In code synthesis, RLEF~\citep{gehring2024rlef} and related work~\citep{le2022coderl,liu2023rltf} use execution feedback to improve generated programs.
These methods establish execution as an effective learning signal.
We focus on the additional learning and credit-assignment problem that arises when the execution result becomes an intermediate state for another action.

\paragraph{Iterative correction and multi-turn policy learning.}
Iterative improvement has been explored through search, self-correction, and multi-turn reinforcement learning~\citep{madaan2023self,shinn2023reflexion,gou2024critic,yao2023react,chen2024teaching}.
DeTikZify refines programs with MCTS, while Sketch2Diagram selects among independently generated candidates.
Recent agentic RL methods study finer-grained credit across interaction steps, including GiGPO~\citep{feng2026group} and turn-level policy optimization~\citep{ding2026empowering}.
Our setting instead uses the execution state produced by the policy's own initial program as the context for a subsequent action, and assigns credit to the initial and corrective actions according to their different roles in the trajectory.

\paragraph{Graphics program synthesis.}
Recent work increasingly treats scientific figure reconstruction as graphics program synthesis rather than raster image generation~\citep{belouadi2025tikzero,rodriguez2025starvector,belouadi2024automatikz,wu2023iconshop}.
DeTikZify~\citep{belouadi2024detikzify} generates TikZ programs from figures and sketches with inference-time refinement, while Sketch2Diagram~\citep{saito2025sketch2diagram} studies sketch-to-TikZ generation with data augmentation and multi-candidate inference.
Omni-I2C~\citep{zhou2026omni} further highlights the difficulty of high-fidelity image-to-code generation across diverse domains.
We use Image-to-TikZ as an executable visual testbed for studying learning after execution.
\section{Conclusion and Limitations}
\label{sec:conclusion}

The study shows that execution can provide more than a terminal learning signal.
It can expose an intermediate state from which additional learning becomes possible.
Using the same training targets, shared policy, and execution score, 
our method turns this state into a second learning interaction and assigns credit according to the distinct roles of initial generation and correction.
This simple change strengthens the policy even under single-call inference, while also producing an additional post-execution corrective capability.
More broadly, our results suggest that executable tasks may contain useful learning structure beyond the scalar reward obtained at the end of generation, 
and that carefully assigning credit can expose this structure without requiring a more complex reward design.
At the same time, our experiments provide an initial demonstration of this principle rather than a complete characterization of its scope.
Our study is limited to one 8B backbone, one executable visual program domain, and a relatively small training set.
Correction is not always beneficial for strong intermediate states, motivating learned stopping or routing, while the instability of absolute corrective credit remains specific to our current RSim-based setting.

\bibliographystyle{unsrt}
\bibliography{references}

\newpage
\appendix

\mainsectionstarnolineno{Appendix}
\addcontentsline{toc}{section}{Appendix}

\noindent\textbf{Appendix Overview}

\noindent
\S\ref{app:qualitative}~~\textbf{Additional Qualitative Results and Failure Cases}
\dotfill \pageref{app:qualitative}

\noindent\hspace{1.5em}
\S\ref{app:qualitative_progression}~~Qualitative Progression across Learning Settings
\dotfill \pageref{app:qualitative_progression}

\noindent\hspace{1.5em}
\S\ref{app:qualitative_revision}~~Successful Visual Revision
\dotfill \pageref{app:qualitative_revision}

\noindent\hspace{1.5em}
\S\ref{app:qualitative_repair}~~Execution Repair
\dotfill \pageref{app:qualitative_repair}

\noindent\hspace{1.5em}
\S\ref{app:qualitative_overrevision}~~Over-Revision of Strong Intermediate States
\dotfill \pageref{app:qualitative_overrevision}

\noindent\hspace{1.5em}
\S\ref{app:qualitative_preservation}~~Preservation and Unresolved Cases
\dotfill \pageref{app:qualitative_preservation}

\noindent
\S\ref{app:workflow_details}~~\textbf{Implementation Details}
\dotfill \pageref{app:workflow_details}

\noindent\hspace{1.5em}
\S\ref{app:workflow_algorithm}~~Full Optimization Algorithm
\dotfill \pageref{app:workflow_algorithm}

\noindent\hspace{1.5em}
\S\ref{app:prompt_contract}~~Prompt and Routing Contract
\dotfill \pageref{app:prompt_contract}

\noindent\hspace{1.5em}
\S\ref{app:routing_details}~~Execution Routing and State Construction
\dotfill \pageref{app:routing_details}

\noindent\hspace{1.5em}
\S\ref{app:training_details}~~Training Configuration
\dotfill \pageref{app:training_details}

\noindent\hspace{1.5em}
\S\ref{app:execution_environment}~~Execution and Reward Environment
\dotfill \pageref{app:execution_environment}

\noindent
\S\ref{app:data_eval}~~\textbf{Data and Evaluation Protocol}
\dotfill \pageref{app:data_eval}

\noindent\hspace{1.5em}
\S\ref{app:training_data}~~Training Data
\dotfill \pageref{app:training_data}

\noindent\hspace{1.5em}
\S\ref{app:heldout_data}~~Held-Out Evaluation Data
\dotfill \pageref{app:heldout_data}

\noindent\hspace{1.5em}
\S\ref{app:evaluation_metrics}~~Evaluation Metrics
\dotfill \pageref{app:evaluation_metrics}

\noindent\hspace{1.5em}
\S\ref{app:inference_protocol}~~Inference and Rendering Protocol
\dotfill \pageref{app:inference_protocol}

\noindent\hspace{1.5em}
\S\ref{app:checkpoint_selection}~~Checkpoint Selection
\dotfill \pageref{app:checkpoint_selection}

\noindent
\S\ref{app:extended_dynamics}~~\textbf{Extended Learning Dynamics and Controls}
\dotfill \pageref{app:extended_dynamics}

\noindent\hspace{1.5em}
\S\ref{app:workflow_full_dynamics}~~Complete Learning Dynamics
\dotfill \pageref{app:workflow_full_dynamics}

\noindent\hspace{1.5em}
\S\ref{app:paired_success_dynamics}~~Paired-Success Correction Dynamics
\dotfill \pageref{app:paired_success_dynamics}

\noindent\hspace{1.5em}
\S\ref{app:continued_direct_controls}~~Continued Direct-GRPO Controls
\dotfill \pageref{app:continued_direct_controls}

\noindent\hspace{1.5em}
\S\ref{app:bootstrap}~~Paired Bootstrap on Final-945
\dotfill \pageref{app:bootstrap}

\noindent
\S\ref{app:credit_analysis}~~\textbf{Credit Assignment Analysis}
\dotfill \pageref{app:credit_analysis}

\noindent\hspace{1.5em}
\S\ref{app:credit_variants}~~Factorized Credit Variants
\dotfill \pageref{app:credit_variants}

\noindent\hspace{1.5em}
\S\ref{app:stage0_credit}~~Downstream versus Local Stage~0 Credit
\dotfill \pageref{app:stage0_credit}

\noindent\hspace{1.5em}
\S\ref{app:credit_preference_audit}~~Direct Audit of Credit-Induced Preferences
\dotfill \pageref{app:credit_preference_audit}

\noindent\hspace{1.5em}
\S\ref{app:credit_stability}~~Optimization Behavior of the Credit Variants
\dotfill \pageref{app:credit_stability}

\noindent\hspace{1.5em}
\S\ref{app:credit_summary}~~Summary
\dotfill \pageref{app:credit_summary}

\noindent
\S\ref{app:workflow_behavior}~~\textbf{State-Dependent Correction and Post-Execution Behavior}
\dotfill \pageref{app:workflow_behavior}

\noindent\hspace{1.5em}
\S\ref{app:state_dependence}~~Full State-Dependent Correction Analysis
\dotfill \pageref{app:state_dependence}

\noindent\hspace{1.5em}
\S\ref{app:state_checkpoint_robustness}~~State Dependence across Checkpoints
\dotfill \pageref{app:state_checkpoint_robustness}

\noindent\hspace{1.5em}
\S\ref{app:routing_behavior}~~Revision and Repair Routing during Training
\dotfill \pageref{app:routing_behavior}

\noindent\hspace{1.5em}
\S\ref{app:repair_analysis}~~Execution Transitions and Repair
\dotfill \pageref{app:repair_analysis}

\noindent\hspace{1.5em}
\S\ref{app:selection_behavior}~~Execution-Aware Selection
\dotfill \pageref{app:selection_behavior}

\noindent\hspace{1.5em}
\S\ref{app:workflow_behavior_summary}~~Summary
\dotfill \pageref{app:workflow_behavior_summary}

\clearpage
\section{Additional Qualitative Results and Failure Cases}
\label{app:qualitative}

\begin{figure}[h]
    \centering
    \includegraphics[width=\linewidth]{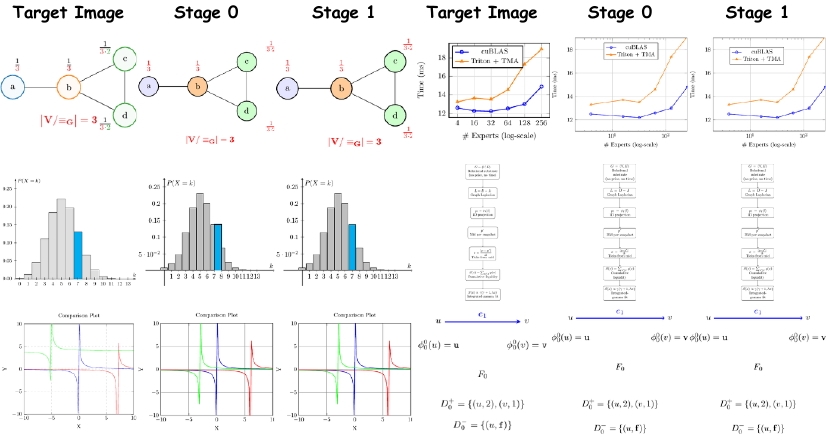}
    \caption{
        Additional qualitative examples of our method from Final-945.
    }
\end{figure}

\begin{figure}[h]
    \centering
    \includegraphics[width=\linewidth]{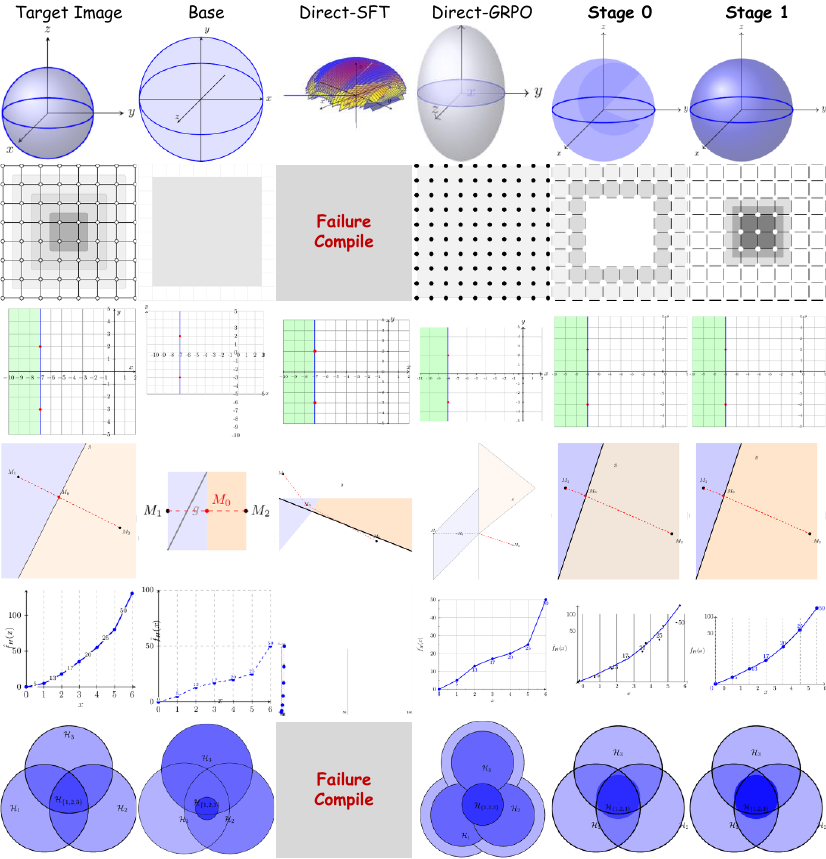}
    \caption{
        Qualitative progression across the main learning settings on Final-945.
        Base model, Direct-SFT, and Direct-GRPO show single-call predictions.
        Ours Stage~0 is also a single-call prediction from the trained policy,
        while Stage~1 is produced after observing the Stage~0 execution state.
        The examples include visible corrective changes and cases where an
        already strong Stage~0 result is largely preserved.
    }
    \label{fig:qualitative_progression}
\end{figure}

This section provides additional qualitative examples from Final-945.
All examples from our method use the checkpoint selected on Dev-150 at
step~100 and the same inference protocol as the quantitative evaluation.
The examples complement the quantitative analyses by illustrating changes
across learning settings, successful visual revision, execution repair,
over-revision, preservation, and unresolved cases.

\subsection{Qualitative Progression across Learning Settings}
\label{app:qualitative_progression}

Figure~\ref{fig:qualitative_progression} compares outputs across the main
learning settings studied in the paper.
For each target image, we show predictions from the base model, Direct-SFT,
Direct-GRPO, and Stage~0 and Stage~1 of our trained policy.
The examples complement the aggregate results in
Table~\ref{tab:final945}.
Execution-feedback training can substantially change the direct prediction
relative to supervised code learning, while our Stage~0 output shows that the
shared policy itself improves further.
Stage~1 then provides an additional opportunity to act on discrepancies
revealed by execution.

Stage~1 does not necessarily introduce a visible change.
When Stage~0 already closely matches the target, the Stage~1 render can remain
effectively unchanged.
These cases are consistent with the state-dependent analysis in
Appendix~\ref{app:state_dependence}, where the marginal value of correction
decreases as the intermediate state becomes stronger.

\subsection{Successful Visual Revision}
\label{app:qualitative_revision}

\begin{figure*}[h]
    \centering
    \includegraphics[width=\linewidth]{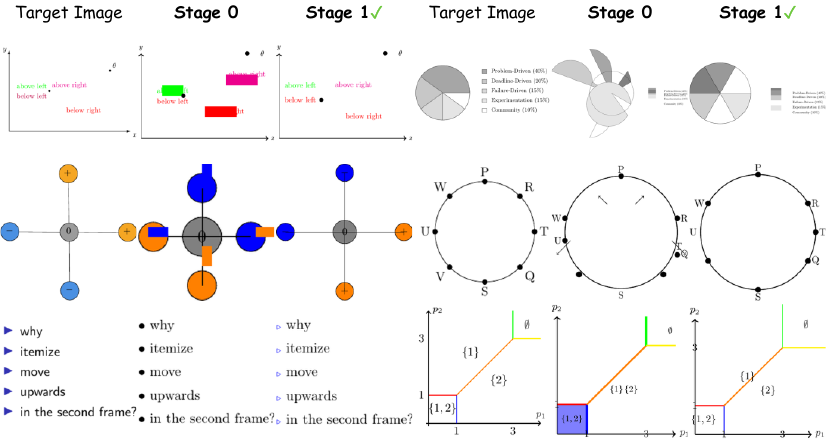}
    \caption{
        Successful visual revisions from Final-945 at checkpoint~100.
        Stage~0 already renders successfully, and Stage~1 revises the executed
        state to better match the target.
        The examples illustrate both large structural corrections and more
        localized changes to geometry, visual elements, or layout.
    }
    \label{fig:qualitative_revision}
\end{figure*}

Figure~\ref{fig:qualitative_revision} shows cases in which Stage~0 already
produces a valid program but its rendered result still differs visibly from the
target.
Stage~1 receives the target image, the Stage~0 render, and the corresponding
TikZ program, and revises the program from this executed state.

The examples show that execution-conditioned correction is not restricted to
small cosmetic edits.
In some cases, Stage~1 substantially reorganizes an incorrect visual
configuration, for example by changing geometric structure or restoring a
missing layout component.
In others, the correction is more localized, such as modifying the shape,
placement, or appearance of a visual element.

The second action is conditioned on the specific program and render produced by
Stage~0 rather than independently resampling from the target image.
The examples therefore illustrate how the policy can use the observed
execution result to modify its own previous program.
The cross-metric improvements reported in
Section~\ref{sec:workflow_correction} and
Appendix~\ref{app:state_dependence} show that these changes are not confined
to RSim.

\subsection{Execution Repair}
\label{app:qualitative_repair}

\begin{figure*}[h]
    \centering
    \includegraphics[width=\linewidth]{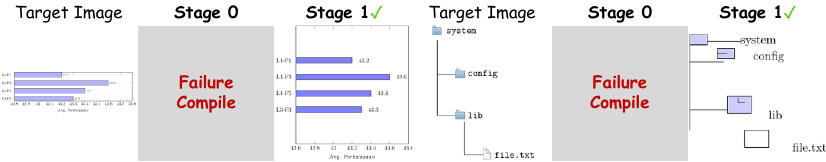}
    \caption{
        Successful execution repair from Final-945.
        Stage~0 does not produce a valid render and therefore receives zero
        execution score.
        Stage~1 conditions on the failed program and its execution status and
        produces a valid TikZ program.
        The illustrated repairs include cases in which Stage~0 has already
        produced a complete document that nevertheless fails compilation.
    }
    \label{fig:qualitative_repair}
\end{figure*}

The corrective action is also used for execution repair.
When Stage~0 fails, no rendered image is available.
Stage~1 instead receives the target image, the unsuccessful TikZ program, and a
compact execution status.
Figure~\ref{fig:qualitative_repair} shows examples in which this route turns an
invalid Stage~0 attempt into a valid program.

The selected examples include complete Stage~0 generations that fail during
compilation, after which Stage~1 modifies the existing program into an
executable one.
The corrective action therefore operates on an unsuccessful program state
rather than simply appending missing tokens.
Repair and visual revision receive different observations but use the same
improvement-relative credit.
For an invalid Stage~0 execution, $S_0=0$, so a successful repair with positive
visual similarity receives positive corrective credit.
Repair trajectories are less frequent than visual-revision trajectories,
as quantified in Appendix~\ref{app:routing_behavior}, while their execution outcomes are analyzed
in Appendix~\ref{app:repair_analysis}.

\subsection{Over-Revision of Strong Intermediate States}
\label{app:qualitative_overrevision}

\begin{figure*}[h]
    \centering
    \includegraphics[width=\linewidth]{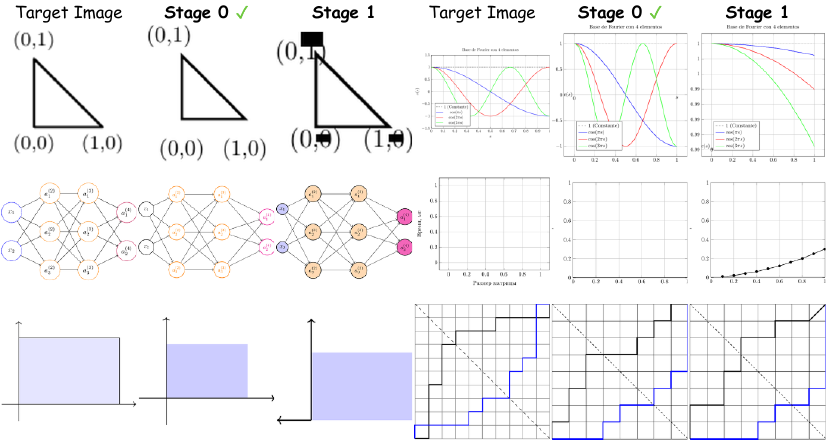}
    \caption{
        Examples of over-revision at checkpoint~100.
        Stage~0 already provides a useful reconstruction, while the Stage~1
        modification reduces RSim or introduces visible artifacts.
        The execution-aware selection rule retains Stage~0 in these cases.
    }
    \label{fig:qualitative_overrevision}
\end{figure*}

Correction is not always beneficial.
Figure~\ref{fig:qualitative_overrevision} shows examples in which Stage~0
already provides a strong reconstruction and Stage~1 modifies the program in a
way that reduces visual fidelity.

Some Stage~1 outputs introduce visible changes that are unsupported by the
target, such as additional graphical elements, distorted annotations, or large
rendering artifacts.
These examples provide a qualitative counterpart to the negative average
correction observed for the strongest Stage~0 bucket in
Appendix~\ref{app:state_dependence}.
Together with the preservation cases in
Figure~\ref{fig:qualitative_progression}, they illustrate the asymmetric value
of the corrective action:
substantial modification can help when the executed state contains clear
discrepancies, while an already strong state may be better left largely
unchanged.

When correction-enabled inference is used, execution-aware selection can retain
Stage~0 when the subsequent correction is worse.
The rule does not generate a new candidate; it only chooses between the two
outputs already produced.

\subsection{Preservation and Unresolved Cases}
\label{app:qualitative_preservation}

\begin{figure*}[h]
    \centering
    \includegraphics[width=\linewidth]{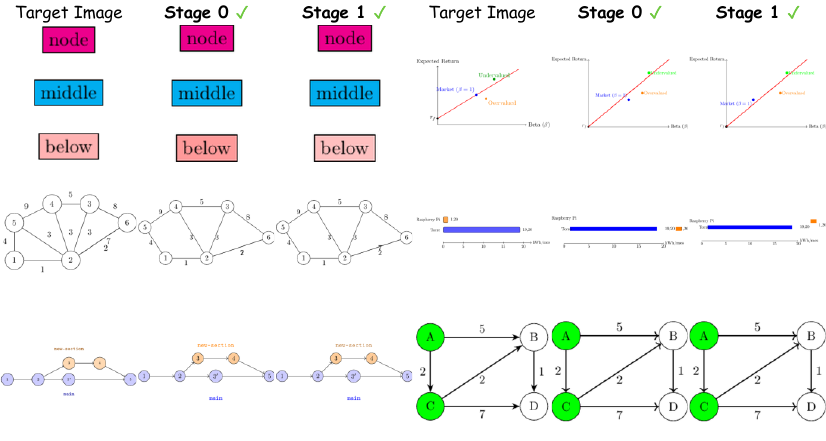}
    \caption{
        Examples of preservation and unresolved cases on Final-945.
        Stage~1 may preserve an already strong Stage~0 result, or leave an
        imperfect reconstruction largely unchanged.
    }
    \label{fig:qualitative_preservation}
\end{figure*}

Stage~1 does not always substantially change the state produced by Stage~0.
Figure~\ref{fig:qualitative_preservation} shows both useful preservation and
cases where important errors remain unresolved.
When Stage~0 already provides a strong reconstruction, preserving the existing
result can avoid unnecessary modification.
In other cases, however, Stage~0 still contains visible errors and Stage~1
changes little.

Some unsuccessful Stage~0 programs also remain non-executable after the repair
action.
Together with the successful revisions, execution repairs, and over-revisions
shown above, these examples illustrate the range of behaviors of the corrective
action: it can improve, degrade, or largely preserve the execution state it
receives.

\section{Implementation Details}
\label{app:workflow_details}

This section provides implementation details omitted from the main text.
We first give the complete optimization procedure, followed by the prompt templates
and execution-dependent routing used for the two actions.
We then summarize the training configuration and execution environment.

\subsection{Full Optimization Algorithm}
\label{app:workflow_algorithm}

Algorithm~\ref{alg:workflow_grpo} summarizes one training iteration of our method.
For each target image, we sample $K=8$ independent complete trajectories.
Each trajectory contains an initial action followed by a corrective action conditioned
on the execution state produced by that initial action.
Both actions are sampled from the same pre-update policy, and no parameter update
occurs during rollout collection.

\begin{algorithm}[h]
\caption{One training iteration of our method}
\label{alg:workflow_grpo}
\begin{algorithmic}[1]
\Require Policy $\pi_{\theta_k}$, target-image batch $\mathcal{B}$,
group size $K=8$, credit scale $\lambda=10$

\For{each target image $I^\star \in \mathcal{B}$}
    \For{$i=1,\ldots,K$}
        \State Sample initial program
        $C_{0,i}\sim\pi_{\theta_k}(\cdot\mid I^\star)$
        \State Execute $C_{0,i}$ and obtain Stage~0 score
        \[
        S_{0,i}=
        \begin{cases}
        \mathrm{RSim}(I_{0,i},I^\star), & \text{if execution succeeds},\\
        0, & \text{otherwise}
        \end{cases}
        \]

        \If{$C_{0,i}$ produces a valid render $I_{0,i}$}
            \State Construct a visual-revision state
            $(I^\star,I_{0,i},C_{0,i})$
            \State Sample
            $C_{1,i}\sim
            \pi_{\theta_k}(\cdot\mid I^\star,I_{0,i},C_{0,i})$
        \Else
            \State Construct a repair state
            $(I^\star,C_{0,i},e_{0,i})$
            \State Sample
            $C_{1,i}\sim
            \pi_{\theta_k}(\cdot\mid I^\star,C_{0,i},e_{0,i})$
        \EndIf

        \State Execute $C_{1,i}$ and obtain
        \[
        S_{1,i}=
        \begin{cases}
        \mathrm{RSim}(I_{1,i},I^\star), & \text{if execution succeeds},\\
        0, & \text{otherwise}
        \end{cases}
        \]

        \State Compute action-specific returns
        \[
        G_{0,i}=S_{1,i},
        \qquad
        G_{1,i}=S_{1,i}-S_{0,i}
        \]
    \EndFor

    \State Mean-center and scale the two return sets independently
    \[
    A_{0,i}=\lambda\left(G_{0,i}-\frac{1}{K}\sum_{j=1}^{K}G_{0,j}\right),
    \qquad
    A_{1,i}=\lambda\left(G_{1,i}-\frac{1}{K}\sum_{j=1}^{K}G_{1,j}\right)
    \]
\EndFor

\State Optimize the Stage~0 loss over action microbatches with weight $0.5$
\State Optimize the Stage~1 loss over action microbatches with weight $0.5$
\end{algorithmic}
\end{algorithm}

Equation~\ref{eq:workflow_loss} in the main text summarizes the relative weighting
of the two action losses.
In the implementation, the two losses are optimized sequentially over action
microbatches rather than accumulated into one optimizer update.
All Stage~0 and Stage~1 trajectories in an iteration are generated before this
optimization begins.

Both Direct-GRPO and our method use the same clipped token-level loss form and
Dr.GRPO-style fixed-length normalization.
For our method, let $\mathcal{M}_{a,b}$ denote action microbatch $b$ of stage
$a\in\{0,1\}$, and let $\theta_{a,b}$ denote the current policy parameters when
that microbatch is processed.
For each completion token, the implementation first computes a detached reference
log probability from the current policy,
\begin{equation}
q^{\mathrm{ref}}_{a,i,t}
=
\operatorname{sg}\!\left[
\log \pi_{\theta_{a,b}}
(c_{a,i,t}\mid x_{a,i},c_{a,i,<t})
\right],
\label{eq:appendix_grpo_reference}
\end{equation}
where $\operatorname{sg}[\cdot]$ denotes stop-gradient and $x_{a,i}$ is the
action-specific context.
A differentiable forward pass of the same current policy is then used to form
\begin{equation}
r_{a,i,t}(\theta)
=
\exp\left(
\log \pi_{\theta}
(c_{a,i,t}\mid x_{a,i},c_{a,i,<t})
-
q^{\mathrm{ref}}_{a,i,t}
\right),
\label{eq:appendix_grpo_ratio}
\end{equation}
with the differentiable forward evaluated at the current parameters
$\theta=\theta_{a,b}$ before the corresponding update.

The clipped token-level surrogate is
\begin{equation}
\ell_{a,i,t}(\theta)
=
-
\min\left[
r_{a,i,t}(\theta) A_{a,i},
\;
\operatorname{clip}
\left(
r_{a,i,t}(\theta),
1-\epsilon_{\mathrm{low}},
1+\epsilon_{\mathrm{high}}
\right)
A_{a,i}
\right].
\label{eq:appendix_grpo_token}
\end{equation}

For an action microbatch $\mathcal{M}_{a,b}$, the optimized loss is
\begin{equation}
\mathcal{L}_{a,b}
=
\frac{w_a}{N_a T_{\max}}
\sum_{i\in\mathcal{M}_{a,b}}
\sum_{t=1}^{T_{\max}}
m_{a,i,t}\,
\ell_{a,i,t}(\theta),
\qquad
w_0=w_1=0.5,
\label{eq:appendix_grpo_loss}
\end{equation}
where $N_a$ is the number of local actions for stage $a$ in the training iteration,
$T_{\max}$ is the maximum completion length, and $m_{a,i,t}$ masks padding and
non-completion positions.
The denominator remains fixed at the stage level when the actions are split into
microbatches.
Completions that reach the generation limit are masked from the policy loss.

We use $\epsilon_{\mathrm{low}}=0.20$ and
$\epsilon_{\mathrm{high}}=0.28$.
No group-standard-deviation normalization is applied, and no KL term is added
($\beta=0$).
Only completion tokens contribute to the policy loss.
Because the detached reference log probabilities are recomputed from the current
policy immediately before each differentiable forward pass, they are not a frozen
copy of the policy that generated the rollout.
Consistent with this implementation, the recorded clipping ratio is zero throughout
the formal 8B run, so the clipping branch is inactive; the advantage-weighted token
loss and its gradients remain active.

For the formal 8B run, training uses four ranks with two target images per rank,
yielding eight target-image groups per training iteration.
Each group contains eight complete trajectories, and each rank receives 16 actions
per stage for optimization.
With an action microbatch size of two, each stage is processed in eight action
microbatches.
Under the executed ZeRO-2 configuration, each such microbatch reaches an AdamW
update boundary, giving eight updates for Stage~0 followed by eight updates for
Stage~1, or 16 AdamW updates per training iteration.
Checkpoint labels such as 25, 50, 75, 100, and 125 therefore refer to training
iterations rather than individual optimizer updates.

\subsection{Prompt and Routing Contract}
\label{app:prompt_contract}

Three prompt templates are used: direct generation, visual revision, and execution repair.
The direct-generation prompt is shared with Direct-SFT and Direct-GRPO and is used for
Stage~0 of our method.
Stage~1 uses either the revision or repair prompt according to the Stage~0 execution result.
The templates below reproduce the prompts used in the formal experiments;
bracketed placeholders denote dynamically inserted state information.

\subsubsection{Direct Generation Prompt}
\label{app:direct_prompt}

The direct prompt receives only the target image.
\begin{figure}[h]
    \centering
    \includegraphics[width=\linewidth]{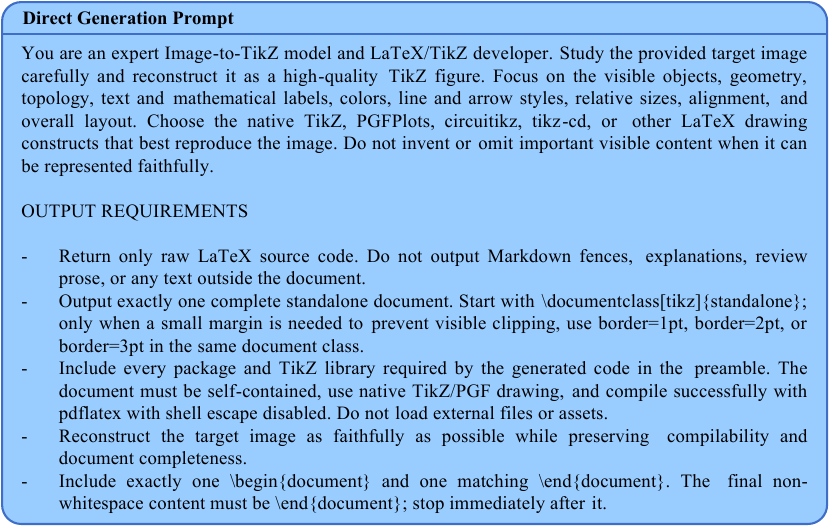}
    \caption{Direct generation prompt.}
    \label{fig:direct_prompt}
\end{figure}

\subsubsection{Visual Revision Prompt}
\label{app:revision_prompt}

When Stage~0 renders successfully, Stage~1 receives the target image, the current
Stage~0 render, and the current TikZ source.
The source is inserted at \texttt{<CURRENT TIKZ PROGRAM>}.

\begin{figure}[h]
    \centering
    \includegraphics[width=\linewidth]{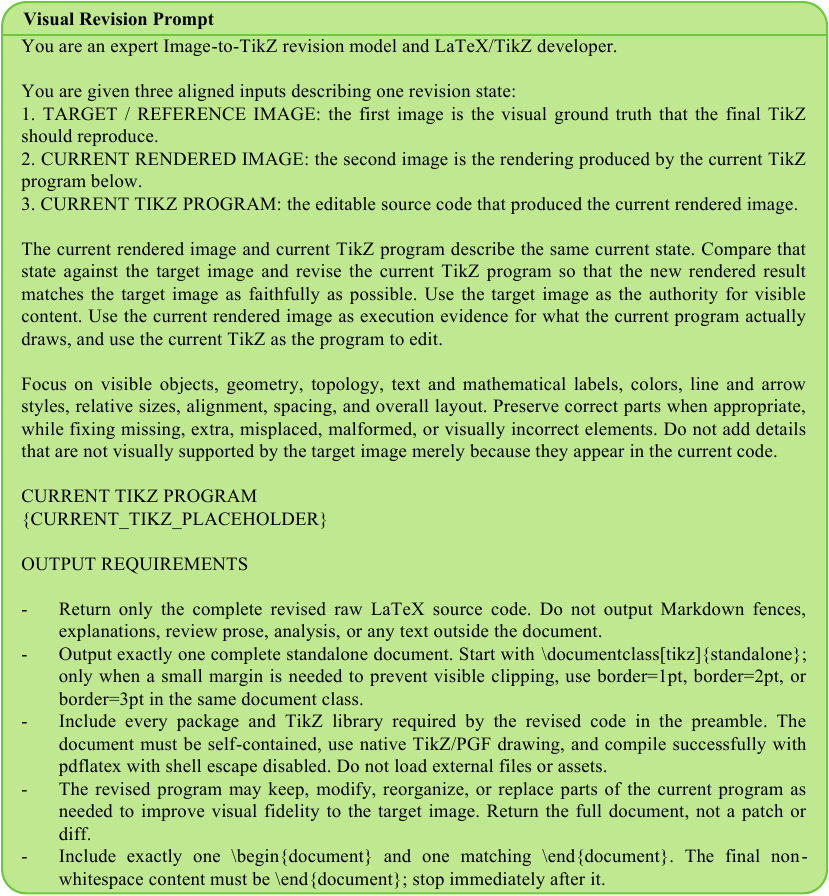}
    \caption{Visual revision prompt.}
    \label{fig:revision_prompt}
\end{figure}

\subsubsection{Execution Repair Prompt}
\label{app:repair_prompt}

When Stage~0 does not produce a valid render, Stage~1 receives the target image,
the unsuccessful TikZ attempt, and a compact execution status instead of a current render.

\begin{figure}[h]
    \centering
    \includegraphics[width=\linewidth]{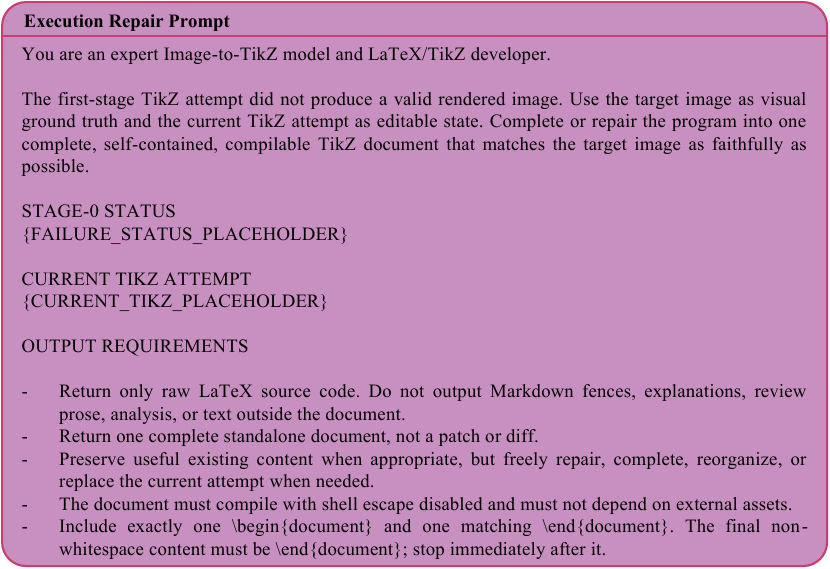}
    \caption{Execution repair prompt.}
    \label{fig:repair_prompt}
\end{figure}

\subsection{Execution Routing and State Construction}
\label{app:routing_details}

Routing is determined solely by the execution result of Stage~0.
A successful render routes the trajectory to visual revision; otherwise it is routed to
repair or completion.
The routing decision does not use RSim or any learned threshold.

For the revision route, the Stage~1 multimodal context is ordered as
\[
\text{target image}
\;\rightarrow\;
\text{current rendered image}
\;\rightarrow\;
\text{current TikZ source}.
\]
The rendered image and TikZ source are produced by the same Stage~0 action, so the second
call observes the visual consequence of the exact program it is asked to modify.

For the repair route, the context contains
\[
\text{target image}
\;\rightarrow\;
\text{compact execution status}
\;\rightarrow\;
\text{current TikZ attempt}.
\]
The status is intentionally compact rather than a full compiler log.
Depending on the failure, it can include the tags
\texttt{hit\_max\_tokens},
\texttt{document\_incomplete},
and the leading category of the renderer/compiler error;
\texttt{render\_failed} is used as a fallback.

A generated program is considered complete at the first
\verb|\end{document}|.
Generation terminates independently for each sequence once this string is produced,
subject to a maximum of 8192 completion tokens for either action.

\subsection{Training Configuration}
\label{app:training_details}

The training subsets and initialization sequence are summarized in
Appendix~\ref{app:training_data}.
Here we report the optimization and runtime configuration used for Direct-SFT,
Direct-GRPO, and our method.

\begin{table}[h]
\caption{
Optimization and runtime configuration for Direct-SFT, Direct-GRPO, and our method.
The final column reports the configuration used by our method in the main experiments.
}
\label{tab:training_config}
\centering
\small
\setlength{\tabcolsep}{4pt}
\begin{tabular}{lccc}
\toprule
& Direct-SFT & Direct-GRPO & Ours \\
\midrule
Training duration
& 3 epochs
& 1 epoch / 125 iterations
& 1 epoch / 125 iterations \\

Learning rate
& $1\times10^{-4}$
& $5\times10^{-6}$
& $5\times10^{-6}$ \\

Weight decay
& --
& $10^{-2}$
& $10^{-2}$ \\

LoRA
& $r=256,\ \alpha=512$
& continued
& continued \\

Group size $K$
& --
& 8
& 8 \\

GRPO loss
& --
& Dr.GRPO
& Dr.GRPO \\

$\beta$
& --
& 0
& 0 \\

Clip range
& --
& $[0.80,1.28]$
& $[0.80,1.28]$ \\

Reward normalization
& --
& none
& none \\

Credit scale $\lambda$
& --
& --
& 10 \\

Action loss weights
& --
& --
& $0.5/0.5$ \\

Sampling temperature
& --
& 1.0
& 1.0 \\

Top-$p$
& --
& 0.9
& 0.9 \\

Maximum completion
& 2,048
& 8,192
& 8,192 / action \\

Image resolution
& $448\times448$
& $448\times448$
& $448\times448$ \\

Precision
& BF16
& BF16
& BF16 \\

Gradient checkpointing
& \checkmark
& \checkmark
& \checkmark \\

Distributed training
& DDP
& ZeRO-2
& ZeRO-2 \\

A100-80GB GPUs
& 2
& 4
& 4 \\
\bottomrule
\end{tabular}
\end{table}

Direct-SFT uses a global batch size of 32
(two GPUs, eight examples per GPU, and two gradient-accumulation steps).
Direct-GRPO uses two target prompts per device with gradient accumulation of eight
and $K=8$ sampled completions per target.
Across four devices, this corresponds to eight target-image groups per training
iteration and one AdamW update per iteration.

Our method also uses two target images per rank on four ranks, yielding eight
target-image groups per training iteration; each group contains eight independent
complete trajectories.
The policy loss is evaluated in action microbatches of two.
Each rank therefore processes 16 actions per stage as eight action microbatches,
which gives 16 AdamW updates per training iteration under the executed ZeRO-2
configuration, as detailed in Appendix~\ref{app:workflow_algorithm}.
The main run uses random seed 20260823.

For our method, rollout generation uses temperature $1.0$, top-$p$ $0.9$,
repetition penalty $1.0$, and a maximum of 8192 tokens for each action.
Stage~0 and Stage~1 generation use microbatch sizes of six and three, respectively.
Training uses AdamW with weight decay $10^{-2}$, a constant learning-rate schedule,
no warmup, and gradient clipping at norm $1.0$.

\subsection{Execution and Reward Environment}
\label{app:execution_environment}

Direct-GRPO and our method use the RSim-based execution score defined in
Eq.~\ref{eq:execution_score} during reinforcement learning.
For our method, execution uses TeX Live 2026 and attempts
\texttt{pdflatex}, \texttt{lualatex}, and \texttt{xelatex}, with shell escape disabled.
Compilation has a 30-second timeout, and rasterization uses 150 DPI before conversion
to the $448\times448$ model input resolution.
Failed compilation or rendering receives $S_t=0$.
The same execution score is used for both actions; their learning signals differ through
the action-specific returns defined in Section~\ref{sec:workflow_credit}.

\section{Data and Evaluation Protocol}
\label{app:data_eval}

This section provides additional details on the training subsets, held-out
evaluation data, evaluation metrics, inference protocol, and checkpoint
selection used throughout the experiments.

\subsection{Training Data}
\label{app:training_data}

All supervised and reinforcement-learning data are drawn from
DaTikZ-V4~\citep{greisinger2026tikzilla}.
Direct-SFT is trained on 3,851 eligible image--TikZ pairs obtained from an
approximately 4K candidate subset after preprocessing.
Direct-GRPO uses 1,000 examples sampled from this supervised-training pool.
Our method is initialized from the final Direct-GRPO policy and reuses the same
1,000 target images.
The corrective action is generated from execution states produced by the
policy's own predictions, so no additional training examples or paired program
annotations are introduced.

\begin{table}[h]
\caption{
Training data used by the main learning settings.
The Direct-GRPO subset is contained in the supervised-training pool, and the
same 1K subset is reused by our method.
}
\label{tab:training_data}
\centering
\small
\begin{tabular}{lcll}
\toprule
Method & Training Examples & Initialization & Training Signal \\
\midrule
Direct-SFT
& 3,851
& Base
& Reference TikZ code \\

Direct-GRPO
& 1,000
& Direct-SFT
& Executed outcome $S$ \\

Ours
& 1,000
& Direct-GRPO
& $S_1$ / $S_1-S_0$ \\
\bottomrule
\end{tabular}
\end{table}

For the continued Direct-GRPO controls in Section~\ref{sec:main_results}, we continue
Direct-GRPO from the same final Direct-GRPO checkpoint on the same 1K target
images while retaining the single-stage Direct-GRPO objective.
We evaluate controls after 100 additional training iterations, matching the
training-iteration budget and target-image exposure at the Dev-selected
checkpoint of our method, and after 125 additional training iterations as an
endpoint comparison.
The controls do not match the number of low-level optimizer updates performed
by our implementation; the corresponding update schedule is described in
Appendix~\ref{app:workflow_details}.

\subsection{Held-Out Evaluation Data}
\label{app:heldout_data}

We construct the evaluation set independently from the training subsets using
the same broad source families as DaTikZ-V4: arXiv, GitHub, and
TeX StackExchange.
Evaluation examples are restricted to source records from 2026 onward; the examples used in this work span January to mid July 2026.
Each retained item contains an executable reference TikZ program and its
corresponding rendered target image, and all examples undergo manual review
during dataset construction.

The paper uses 1,095 held-out examples.
We fix 150 examples as \textbf{Dev-150} for checkpoint selection, ablations,
and behavioral analysis, and reserve the remaining 945 examples as
\textbf{Final-945} for final evaluation.
Final-945 is not used for checkpoint selection or method development.

\begin{table}[h]
    \caption{
        Composition of the held-out evaluation set.
        All samples are drawn from source records dated in 2026.
        }
    \label{tab:heldout_sources}
    \centering
    \small
    \setlength{\tabcolsep}{10pt}
    \begin{tabular}{lcccc}
    \toprule
    Split & arXiv & GitHub & TeX.SE & Total \\
    \midrule
    Dev-150   & 82  & 43  & 25  & 150 \\
    Final-945 & 458 & 309 & 178 & 945 \\
    \midrule
    All       & 540 & 352 & 203 & 1,095 \\
    \bottomrule
    \end{tabular}
\end{table}

The two splits contain distinct evaluation examples and serve different
purposes throughout the paper:
all checkpoint-dependent analysis is performed on Dev-150, while Final-945 is
evaluated only after the checkpoint and inference procedure have been fixed.

\subsection{Evaluation Metrics}
\label{app:evaluation_metrics}

We evaluate both executability and visual fidelity.
\emph{Render success} indicates whether a generated program can be successfully
compiled and rasterized.
For a valid render $I$, visual fidelity to the target image $I^\star$ is
measured using four complementary image-similarity metrics:
RSim~\citep{greisinger2026tikzilla},
DINO~\citep{caron2021emerging},
DreamSim~\citep{fu2023DreamSim},
and CLIPImg~\citep{radford2021learning}.

RSim uses the DeTikZify-V2 vision encoder and serves as the only visual
optimization reward in Direct-GRPO and our method.
DINO, DreamSim, and CLIPImg are used only for evaluation and do not participate
in training or execution-aware selection.
We report DreamSim similarity as one minus the perceptual distance returned by
the official DreamSim implementation.

\begin{table}[h]
    \caption{
        Evaluation metrics and their roles.
        All main visual results use the fixed-zero convention.
        }
    \label{tab:metric_roles}
    \centering
    \small
    \begin{tabular}{lll}
    \toprule
    Metric & Role in Training & Main Evaluation Convention \\
    \midrule
    Render success & Execution signal & Success / failure \\
    RSim            & RL reward + evaluation & Failure $\rightarrow 0$ \\
    DINO            & Evaluation only & Failure $\rightarrow 0$ \\
    DreamSim        & Evaluation only & Failure $\rightarrow 0$ \\
    CLIPImg         & Evaluation only & Failure $\rightarrow 0$ \\
    \bottomrule
    \end{tabular}
\end{table}

Under the \emph{fixed-zero} convention, unsuccessful executions receive zero
for every visual metric and successful executions retain their measured
similarity.
This convention is used for the main aggregate results because it reflects both
executability and visual quality.
When analyzing the visual effect of the corrective action independently of
execution recovery, we instead restrict to the \emph{paired-success} subset in
which both Stage~0 and Stage~1 render successfully, as defined in
Section~\ref{sec:workflow_correction}.

\subsection{Inference and Rendering Protocol}
\label{app:inference_protocol}

All paper-active evaluations use stochastic decoding with temperature $1.0$,
top-$p$ $0.9$, repetition penalty $1.0$, a maximum of 8192 newly generated
tokens, and early termination at the first complete
\verb|\end{document}|.

Direct models produce a single output from the target image.
For our method, Stage~0 is generated from the target image alone and is reported
as the single-call inference output.
In correction-enabled evaluation, the same policy is called again after
Stage~0 execution to produce Stage~1 according to the routing procedure in
Appendix~\ref{app:routing_details}.

The final \emph{Selected} output is chosen only from the two programs already
produced:
\begin{itemize}
    \item if only Stage~0 renders, Stage~0 is retained;
    \item if only Stage~1 renders, Stage~1 is retained;
    \item if both render, the program with higher RSim is retained;
    \item ties are assigned to Stage~0;
    \item if neither renders, the selected output is unsuccessful.
\end{itemize}
No additional generation is performed by the selection procedure.

Generation parameters and evaluation metrics are unified across the
paper-active evaluations.
The Direct Final-945 evaluation uses a 120-second compilation timeout, whereas
the correction-enabled renderer uses a 30-second timeout; all successful Direct
predictions in Final-945 execute within 30 seconds.
Both pipelines normalize rendered images to $448\times448$ before visual
evaluation.

\subsection{Checkpoint Selection}
\label{app:checkpoint_selection}

Checkpoints of our method are evaluated on Dev-150 at the initializer and after
25, 50, 75, 100, and 125 training iterations.
For consistency with the training logs and figures, we retain the checkpoint
labels Step~25 through Step~125 throughout the paper.
For each checkpoint, we evaluate Stage~0, raw Stage~1, and the
execution-aware selected output using the same inference and metric protocol.

Table~\ref{tab:checkpoint_selection} reports the quantities used to compare the
checkpoints.
Among the evaluated checkpoints, checkpoint~100 achieves the highest selected render
rate as well as the highest selected RSim, DINO, DreamSim, and CLIPImg scores.
We therefore fix checkpoint~100 for subsequent single-policy behavioral
analyses and for the final evaluation on Final-945.
Training-dynamics analyses instead report the full checkpoint trajectory.

\begin{table*}[h]
\caption{
Checkpoint evaluation for our method on Dev-150.
Visual metrics are computed on the execution-aware selected output using the
fixed-zero convention.
Checkpoint~100 gives the highest selected render rate and all four selected
visual metrics.
}
\label{tab:checkpoint_selection}
\centering
\small
\setlength{\tabcolsep}{4pt}
\begin{tabular}{lcccccccc}
\toprule
Checkpoint
& Stage~0 Render
& Stage~1 Render
& Selected Render
& RSim
& DINO
& DreamSim
& CLIPImg \\
\midrule
Initializer
& 82.00\%
& 82.67\%
& 82.67\%
& 0.4736
& 0.7088
& 0.6497
& 0.7468 \\

Step 25
& 87.33\%
& 88.00\%
& 89.33\%
& 0.5161
& 0.7667
& 0.7076
& 0.8116 \\

Step 50
& 89.33\%
& 88.00\%
& 90.67\%
& 0.5223
& 0.7805
& 0.7160
& 0.8196 \\

Step 75
& 91.33\%
& 90.00\%
& 92.00\%
& 0.5454
& 0.7975
& 0.7386
& 0.8416 \\

\textbf{Step 100}
& \textbf{96.67\%}
& \textbf{94.67\%}
& \textbf{97.33\%}
& \textbf{0.5646}
& \textbf{0.8326}
& \textbf{0.7677}
& \textbf{0.8774} \\

Step 125
& 95.33\%
& 92.67\%
& 95.33\%
& 0.5549
& 0.8117
& 0.7538
& 0.8666 \\
\bottomrule
\end{tabular}
\end{table*}

The selected checkpoint is fixed before Final-945 evaluation.
Final-945 is therefore used only to report final performance rather than to
choose between training checkpoints or analysis settings.

\section{Extended Learning Dynamics and Controls}
\label{app:extended_dynamics}

This section provides the complete numerical results behind the learning
dynamics reported in the main text, compares our method with continued
single-stage Direct-GRPO under matched training-iteration budgets, and quantifies sample-level
uncertainty for selected Final-945 comparisons.
All Dev-150 checkpoint results follow the evaluation protocol described in
Appendix~\ref{app:data_eval}.

\subsection{Complete Learning Dynamics}
\label{app:workflow_full_dynamics}

We evaluate the initializer and checkpoints at steps 25, 50, 75, 100, and 125 on Dev-150.
Table~\ref{tab:workflow_execution_dynamics} reports the execution behavior of
both actions together with the four possible execution transitions.
We denote successful and failed executions by $\mathrm{S}$ and $\mathrm{F}$,
respectively.
For example, $\mathrm{FS}$ denotes a Stage~0 failure followed by a successful
Stage~1 repair.

\begin{table}[h]
\caption{
Complete execution dynamics of our method on Dev-150.
Each row contains 150 examples.
$\mathrm{SS}$, $\mathrm{SF}$, $\mathrm{FS}$, and $\mathrm{FF}$ denote the
execution outcomes of Stage~0 and Stage~1, respectively.
}
\label{tab:workflow_execution_dynamics}
\centering
\small
\setlength{\tabcolsep}{5pt}
\begin{tabular}{lcccccc}
\toprule
Checkpoint
& Stage~0 Render
& Stage~1 Render
& SS
& SF
& FS
& FF \\
\midrule
Initializer
& 82.00\%
& 82.67\%
& 123 & 0 & 1 & 26 \\

Step 25
& 87.33\%
& 88.00\%
& 129 & 2 & 3 & 16 \\

Step 50
& 89.33\%
& 88.00\%
& 130 & 4 & 2 & 14 \\

Step 75
& 91.33\%
& 90.00\%
& 134 & 3 & 1 & 12 \\

Step 100
& 96.67\%
& 94.67\%
& 141 & 4 & 1 & 4 \\

Step 125
& 95.33\%
& 92.67\%
& 139 & 4 & 0 & 7 \\
\bottomrule
\end{tabular}
\end{table}

The execution rates improve substantially over training.
Stage~0 rises from $82.00\%$ for the initializer to $96.67\%$ at step 100,
while Stage~1 rises from $82.67\%$ to $94.67\%$.
Most of this change comes from the reduction of $\mathrm{FF}$ trajectories.
As Stage~0 improves, failed initial executions also become less frequent, so
the repair route is exercised less often on later Dev-150 checkpoints.
Appendix~\ref{app:workflow_behavior} analyzes routing and repair in more detail.

\subsection{Paired-Success Correction Dynamics}
\label{app:paired_success_dynamics}

To isolate visual correction from execution recovery, we use the paired-success
subset defined in Section~\ref{sec:workflow_correction}.
For each visual metric $M$, the reported quantity is
\[
\Delta M = M(I_1,I^\star)-M(I_0,I^\star).
\]
Positive values indicate that Stage~1 improves the already executable Stage~0
state.

Table~\ref{tab:workflow_visual_dynamics} reports the complete paired-success
visual dynamics.
The correction-outcome categories follow the definition used in the main text:
$\Delta\mathrm{RSim}>0.005$ is an improvement,
$|\Delta\mathrm{RSim}|\leq0.005$ is neutral,
and $\Delta\mathrm{RSim}<-0.005$ is a degradation.

\begin{table*}[h]
\caption{
Visual correction dynamics on the paired-success subset of Dev-150.
The four $\Delta$ columns report the mean Stage~1 minus Stage~0 change.
The last three columns show the composition of correction outcomes under the
$\pm0.005$ RSim neutral region used in the main text.
}
\label{tab:workflow_visual_dynamics}
\centering
\small
\setlength{\tabcolsep}{4.5pt}
\begin{tabular}{lrrrrrrrr}
\toprule
Checkpoint
& Paired $n$
& $\Delta$RSim
& $\Delta$DINO
& $\Delta$DreamSim
& $\Delta$CLIPImg
& Improve
& Neutral
& Degrade \\
\midrule
Initializer
& 123
& $-0.0031$
& $-0.0027$
& $-0.0039$
& $-0.0017$
& 12.20\%
& 65.04\%
& 22.76\% \\

Step 25
& 129
& $+0.0013$
& $-0.0014$
& $+0.0043$
& $+0.0042$
& 28.68\%
& 48.06\%
& 23.26\% \\

Step 50
& 130
& $+0.0098$
& $+0.0053$
& $+0.0077$
& $+0.0016$
& 43.85\%
& 26.92\%
& 29.23\% \\

Step 75
& 134
& $+0.0121$
& $+0.0058$
& $+0.0112$
& $+0.0063$
& 42.54\%
& 30.60\%
& 26.87\% \\

Step 100
& 141
& $-0.0012$
& $-0.0072$
& $-0.0047$
& $-0.0000$
& 43.26\%
& 19.86\%
& 36.88\% \\

Step 125
& 139
& $+0.0065$
& $+0.0057$
& $+0.0025$
& $-0.0013$
& 46.76\%
& 19.42\%
& 33.81\% \\
\bottomrule
\end{tabular}
\end{table*}

The complete trajectory highlights the distinction between initial generation
quality and corrective behavior.
The initializer has negative average Stage~1 changes under all four visual
metrics.
By steps 50--75, correction becomes more effective, and all four mean changes
are positive at step 75.
The trend is not monotonic: at step 100, Stage~0 reaches its highest render
rate, while the average paired-success change becomes slightly negative.
At step 125, RSim, DINO, and DreamSim again show positive average changes.

Checkpoint selection is therefore based on the complete inference result rather
than the paired-success correction gain alone.
Step 100 provides the strongest selected result on Dev-150
(Appendix~\ref{app:checkpoint_selection}), while
Appendix~\ref{app:state_dependence} shows that its corrective behavior varies
with the quality of the intermediate state.

\subsection{Continued Direct-GRPO Controls}
\label{app:continued_direct_controls}

We compare our method with continued single-stage Direct-GRPO from the same
Direct-GRPO initializer and on the same 1K training images.
The $+100$ comparison matches the additional training-iteration budget and
target-image exposure at the Dev-selected checkpoint used for final evaluation.
The $+125$ comparison uses the corresponding training endpoint.
The controls match initialization, training data, training-iteration count, and
target-image exposure, but not the number of low-level optimizer updates; our
method additionally samples the execution-conditioned corrective action during
training.

Table~\ref{tab:continued_direct_single_call} compares single-call outputs on
Final-945.
For our method, this is the Stage~0 output generated directly from the target
image without invoking correction at inference time.

\begin{table*}[h]
\caption{
Continued Direct-GRPO single-call results on Final-945.
Both methods start from the same Direct-GRPO checkpoint and use the same 1K
training images.
The $+100$ row corresponds to the Dev-selected checkpoint of our method;
$+125$ reports the training endpoint.
}
\label{tab:continued_direct_single_call}
\centering
\small
\setlength{\tabcolsep}{5.5pt}
\begin{tabular}{llccccc}
\toprule
Additional Iterations
& Method
& Render
& RSim
& DINO
& DreamSim
& CLIPImg \\
\midrule

$+100$
& Continued Direct-GRPO
& 88.25\%
& 0.5135
& 0.7629
& 0.7019
& 0.7996 \\

$+100$
& Ours (Stage~0)
& \textbf{92.70\%}
& \textbf{0.5382}
& \textbf{0.8014}
& \textbf{0.7330}
& \textbf{0.8372} \\

\midrule

$+125$
& Continued Direct-GRPO
& 87.20\%
& 0.5119
& 0.7557
& 0.6984
& 0.7952 \\

$+125$
& Ours (Stage~0)
& \textbf{93.54\%}
& \textbf{0.5356}
& \textbf{0.8011}
& \textbf{0.7346}
& \textbf{0.8431} \\

\bottomrule
\end{tabular}
\end{table*}

Continued Direct-GRPO improves over the original Direct-GRPO checkpoint, but
our Stage~0 output is higher under render success and all four visual metrics
at both corresponding training endpoints.
The gain from training is therefore already present in the initial action and
does not require the second call at inference time.

We also evaluate the continued Direct-GRPO policies with the same
execution-conditioned second call.
Table~\ref{tab:continued_direct_correction} reports paired-success visual
changes on Final-945.
At both corresponding endpoints, the continued single-stage policies produce changes
close to zero, whereas our corrective action yields larger positive mean
changes across the reported metrics.

\begin{table*}[h]
\caption{
Continued Direct-GRPO paired-success correction on Final-945.
$\Delta M$ denotes Stage~1 minus Stage~0 on examples where both outputs render
successfully.
}
\label{tab:continued_direct_correction}
\centering
\small
\setlength{\tabcolsep}{5.5pt}
\begin{tabular}{llccccc}
\toprule
\shortstack{Additional\\Iterations}
& Method
& Paired $n$
& $\Delta$RSim
& $\Delta$DINO
& $\Delta$DreamSim
& $\Delta$CLIPImg \\
\midrule

$+100$
& Continued Direct-GRPO
& 828
& $+0.0002$
& $+0.0019$
& $+0.0020$
& $-0.0007$ \\

$+100$
& Ours
& 863
& $+0.0073$
& $+0.0034$
& $+0.0074$
& $+0.0031$ \\

\midrule

$+125$
& Continued Direct-GRPO
& 815
& $+0.0017$
& $+0.0013$
& $+0.0009$
& $+0.0008$ \\

$+125$
& Ours
& 869
& $+0.0079$
& $+0.0046$
& $+0.0068$
& $+0.0035$ \\

\bottomrule
\end{tabular}
\end{table*}

For the $+125$ continued Direct-GRPO correction analysis, paired statistics use
the Stage~0 records from the correction-enabled evaluation run.
The canonical single-call Direct result is reported separately in
Table~\ref{tab:continued_direct_single_call}.

\subsection{Paired Bootstrap on Final-945}
\label{app:bootstrap}

We quantify uncertainty over the fixed Final-945 evaluation set using a paired
bootstrap.
For each comparison, we resample the same 945 example indices with replacement
for both methods and compute the difference in their mean performance.
We use 10,000 paired resamples with a fixed random seed of 20260914 and report
percentile $95\%$ intervals.
Positive differences indicate better performance for the second method.

Table~\ref{tab:final945_bootstrap} reports three fixed comparisons:
from supervised learning to direct execution feedback,
from Direct-GRPO to the Stage~1 output of our trained policy,
and from raw Stage~1 to execution-aware selection.

\begin{table}[h]
    \caption{
        Paired bootstrap uncertainty on the fixed Final-945 evaluation set
        ($N=945$, 10,000 paired resamples).
        Each entry reports the observed difference with its $95\%$ bootstrap interval.
        Positive values favor the second method.
        Render differences are in percentage points; visual metrics use fixed-zero scores.
        Intervals quantify uncertainty over the fixed evaluation samples rather than variation across training seeds.
        }
    \label{tab:final945_bootstrap}
    \centering
    \small
    \setlength{\tabcolsep}{6pt}
    \renewcommand{\arraystretch}{1.15}

    \begin{tabular}{lccc}
    \toprule
    Metric
    & \shortstack{SFT $\rightarrow$ Direct-GRPO}
    & \shortstack{Direct-GRPO $\rightarrow$ Ours Stage~1}
    & \shortstack{Stage~1 $\rightarrow$ Selected} \\
    \midrule

    $\Delta$ Render Success
    & \shortstack{$+17.25$ pp\\{\scriptsize $[+13.86,+20.63]$}}
    & \shortstack{$+10.90$ pp\\{\scriptsize $[+8.25,+13.54]$}}
    & \shortstack{$+1.38$ pp\\{\scriptsize $[+0.74,+2.12]$}}
    \\

    $\Delta$ RSim
    & \shortstack{$+0.1154$\\{\scriptsize $[+0.0962,+0.1345]$}}
    & \shortstack{$+0.0738$\\{\scriptsize $[+0.0580,+0.0892]$}}
    & \shortstack{$+0.0203$\\{\scriptsize $[+0.0162,+0.0249]$}}
    \\

    $\Delta$ DINO
    & \shortstack{$+0.1619$\\{\scriptsize $[+0.1334,+0.1905]$}}
    & \shortstack{$+0.1000$\\{\scriptsize $[+0.0765,+0.1238]$}}
    & \shortstack{$+0.0175$\\{\scriptsize $[+0.0113,+0.0243]$}}
    \\

    $\Delta$ DreamSim
    & \shortstack{$+0.1508$\\{\scriptsize $[+0.1244,+0.1773]$}}
    & \shortstack{$+0.0932$\\{\scriptsize $[+0.0721,+0.1149]$}}
    & \shortstack{$+0.0192$\\{\scriptsize $[+0.0137,+0.0254]$}}
    \\

    $\Delta$ CLIPImg
    & \shortstack{$+0.1629$\\{\scriptsize $[+0.1333,+0.1939]$}}
    & \shortstack{$+0.0999$\\{\scriptsize $[+0.0755,+0.1245]$}}
    & \shortstack{$+0.0174$\\{\scriptsize $[+0.0111,+0.0246]$}}
    \\

    \bottomrule
    \end{tabular}
\end{table}

The intervals exclude zero for all five metrics in each comparison.
Direct execution feedback provides a large improvement over supervised code
learning, the trained policy provides a further positive shift over the
preceding Direct-GRPO checkpoint, and execution-aware selection adds a smaller
but consistent improvement over raw Stage~1.

These intervals quantify uncertainty over the fixed Final-945 sample set.
They do not measure variation across independent training runs.

\section{Credit Assignment Analysis}
\label{app:credit_analysis}

This section expands the credit-assignment analysis in
Section~\ref{sec:credit_ablation}.
We first report the controlled comparison between downstream and local credit
for the initial action.
We then compare the group-relative preferences induced by alternative credit
definitions on the same sampled trajectories, followed by the optimization
behavior of the four factorized variants.

\subsection{Factorized Credit Variants}
\label{app:credit_variants}

The ablation varies the credit assigned to the initial and corrective actions.
Stage~0 either receives the final trajectory score $S_1$ or its local
intermediate score $S_0$.
Stage~1 either receives its improvement over the intermediate state,
$S_1-S_0$, or the absolute final score $S_1$.
All variants use the same group centering and scale $\lambda=10$ defined in
Section~\ref{sec:workflow_optimization}.

\begin{table}[h]
\caption{
Factorized credit-assignment variants.
Variant A is our method.
The table reports the raw credit before group centering and scaling.
}
\label{tab:appendix_credit_variants}
\centering
\small
\setlength{\tabcolsep}{8pt}
\begin{tabular}{lcc}
\toprule
Variant
& Stage~0 credit
& Stage~1 credit \\
\midrule
A (Ours)
& $S_1$
& $S_1-S_0$ \\

B
& $S_0$
& $S_1-S_0$ \\

C
& $S_1$
& $S_1$ \\

D
& $S_0$
& $S_1$ \\
\bottomrule
\end{tabular}
\end{table}

A versus B isolates the Stage~0 credit while keeping improvement-relative
Stage~1 credit fixed.
A versus C isolates the Stage~1 credit while keeping downstream Stage~0 credit
fixed.
Variants C and D both use absolute Stage~1 credit with different Stage~0
definitions.

\subsection{Downstream versus Local Stage~0 Credit}
\label{app:stage0_credit}

Table~\ref{tab:ab_full} gives the complete Dev-150 comparison between variants
A and B at the checkpoints used in the main analysis.
The paired visual changes are computed only on examples for which both stages
render successfully.

\begin{table*}[h]
\caption{
Comparison of downstream Stage~0 credit (A) and local Stage~0 credit (B) on
Dev-150.
$\Delta M$ denotes the mean Stage~1 minus Stage~0 change on the paired-success
subset.
A and B use the same improvement-relative Stage~1 credit and differ only in
whether Stage~0 receives $S_1$ or $S_0$.
}
\label{tab:ab_full}
\centering
\small
\setlength{\tabcolsep}{4pt}
\begin{tabular}{llcccccc}
\toprule
Checkpoint
& Variant
& Stage~0 Render
& Stage~1 Render
& $\Delta$RSim
& $\Delta$DINO
& $\Delta$DreamSim
& $\Delta$CLIPImg \\
\midrule

Step 25
& A
& 87.33\%
& 88.00\%
& $+0.0013$
& $-0.0014$
& $+0.0043$
& $+0.0042$ \\

& B
& 84.67\%
& 86.00\%
& $+0.0012$
& $-0.0015$
& $+0.0005$
& $-0.0007$ \\
\midrule

Step 50
& A
& 89.33\%
& 88.00\%
& $+0.0098$
& $+0.0053$
& $+0.0077$
& $+0.0016$ \\

& B
& 90.00\%
& 89.33\%
& $+0.0002$
& $+0.0023$
& $+0.0038$
& $+0.0001$ \\
\midrule

Step 75
& A
& 91.33\%
& 90.00\%
& $+0.0121$
& $+0.0058$
& $+0.0112$
& $+0.0063$ \\

& B
& 88.00\%
& 88.67\%
& $-0.0008$
& $-0.0023$
& $-0.0037$
& $-0.0014$ \\
\bottomrule
\end{tabular}
\end{table*}

The comparison separates Stage~0 quality from the behavior of the subsequent
corrective action.
At step 50, B has a slightly higher Stage~0 render rate than A
($90.00\%$ versus $89.33\%$), and its Stage~0 RSim-FZ is also higher
($0.5095$ versus $0.4960$).
Its paired Stage~1 correction, however, is nearly neutral:
$\Delta\mathrm{RSim}=+0.0002$, compared with $+0.0098$ for A.
By step 75, A retains a positive correction gain of $+0.0121$, whereas B
becomes slightly negative at $-0.0008$.

Thus, a stronger intermediate output at one checkpoint does not imply a
stronger subsequent correction.
Using $S_1$ rather than $S_0$ for Stage~0 changes which initial actions are
favored according to the quality reached after the corrective action.

\subsection{Direct Audit of Credit-Induced Preferences}
\label{app:credit_preference_audit}

The trained ablations compare the behavior produced by different credit
definitions.
We additionally inspect the credit signals while holding the sampled
trajectories fixed.
For this analysis, we use the completed A training run, containing 1,000
target-image groups and eight independently sampled complete trajectories per
group, for a total of 8,000 trajectories.

For Stage~0, we compare the downstream advantage
\[
A_0(S_1)
=
S_1-\overline{S_1}
\]
with the local advantage
\[
A_0(S_0)
=
S_0-\overline{S_0}.
\]
For Stage~1, we compare the improvement-relative advantage
\[
A_1(S_1-S_0)
=
(S_1-S_0)-\overline{(S_1-S_0)}
\]
with the absolute-score advantage
\[
A_1(S_1)
=
S_1-\overline{S_1}.
\]
The common factor $\lambda=10$ is omitted because it does not affect
correlation, sign, or ranking.

\begin{table*}[h]
    \caption{
        Comparison of group-relative advantages induced by alternative credit
        definitions on the same 8,000 trajectories from 1,000 target-image
        groups.
        Sign flips are computed over all trajectories, with exact-zero
        advantages counted as neither positive nor negative.
        Top-1 change denotes the fraction of groups whose highest-advantage
        trajectory changes under the alternative definition.
        }
    \label{tab:credit_preference_audit}
    \centering
    \small
    \setlength{\tabcolsep}{5pt}
    \renewcommand{\arraystretch}{1.08}

    \begin{tabular}{lccccc}
    \toprule
    Advantage Comparison
    & Pearson
    & Spearman
    & \shortstack{Sign Flip}
    & \shortstack{Group Top-1 Change}
    & \shortstack{Pairwise Ordering\\Change} \\
    \midrule

    $A_0(S_1)$ vs.\ $A_0(S_0)$
    & 0.7408
    & 0.7083
    & 21.4\%
    & 46.1\%
    & -- \\

    $A_1(S_1-S_0)$ vs.\ $A_1(S_1)$
    & 0.3848
    & 0.3146
    & 37.2\%
    & 74.5\%
    & 41.76\% \\

    \bottomrule
    \end{tabular}
\end{table*}

For Stage~0, the two definitions remain moderately correlated but produce
different group-relative preferences.
The advantage sign changes for
$1{,}715/8{,}000=21.4\%$ of trajectories, and the highest-advantage trajectory
changes in $461/1{,}000=46.1\%$ of target-image groups.

The distinction is larger for Stage~1.
Improvement-relative and absolute credit have a Pearson correlation of
$0.3848$ and a Spearman correlation of $0.3146$.
The advantage sign changes for
$2{,}974/8{,}000=37.2\%$ of trajectories, and the highest-advantage corrective
trajectory changes in $745/1{,}000=74.5\%$ of groups.

We additionally compare every unordered pair of trajectories within each group.
For a pair $(i,j)$, we compare the sign of
\[
A_{1,i}(S_1-S_0)-A_{1,j}(S_1-S_0)
\]
with the sign of
\[
A_{1,i}(S_1)-A_{1,j}(S_1).
\]
Ties form a separate ordering category.
Of the $28{,}000$ within-group trajectory pairs,
$11{,}693$ change their relative ordering, corresponding to $41.76\%$.
Thus, $S_1-S_0$ and $S_1$ do not merely rescale the same corrective preference;
they change the relative ordering of corrective actions within a group.

\subsection{Optimization Behavior of the Credit Variants}
\label{app:credit_stability}

The fixed-trajectory audit above compares the learning signals directly.
The corresponding training runs also differ in optimization behavior.
Table~\ref{tab:credit_stability} summarizes the four variants.

\begin{table}[h]
\caption{
Optimization behavior observed for the four factorized credit variants.
}
\label{tab:credit_stability}
\centering
\small
\setlength{\tabcolsep}{8pt}
\begin{tabular}{lccl}
\toprule
Variant
& Stable Until
& First Abnormal Step
& Observed Behavior \\
\midrule

A
& 125
& --
& Completed normally \\

B
& $\sim81$
& 82
& Late length/termination pathology \\

C
& 24
& 25--26
& Early zero-signal collapse \\

D
& 25
& 26
& Early zero-signal collapse \\
\bottomrule
\end{tabular}
\end{table}

Variant A completes all 125 training iterations without the length or termination
failure observed in the alternative runs.
Variant B behaves normally through the checkpoints used for the A--B comparison
but develops a late-training length pathology.
Its mean Stage~1 completion length is 482 tokens at step 80,
965 tokens at step 81, and 3,663 tokens at step 82.
At step 83 it increases to 8,067 tokens, with $98.4\%$ of Stage~1 completions
reaching the generation limit and no Stage~1 completion terminating normally at
\verb|\end{document}|.

Variants C and D show an earlier shared failure regime.
Both use absolute Stage~1 credit $S_1$ but different Stage~0 credits.
Their execution behavior collapses around step 26.
Table~\ref{tab:cd_collapse} shows the transition around this point.

\begin{table*}[h]
    \caption{
        Training diagnostics around the C/D collapse.
        ``Hit Max'' is the fraction of Stage~1 generations reaching the token
        limit, ``End Doc'' is the fraction terminating normally at
        \texttt{\textbackslash end\{document\}}, and $A_0/A_1$ nonzero counts
        the trajectories receiving nonzero group-relative advantages.
        Each training iteration contains 64 complete trajectories.
        }
    \label{tab:cd_collapse}
    \centering
    \small
    \setlength{\tabcolsep}{6pt}
    \begin{tabular}{llcccccccc}
    \toprule
    \shortstack{Variant}
    & Step
    & \shortstack{Stage~1\\Tokens}
    & Hit Max
    & End Doc
    & \shortstack{Stage~0\\Compile}
    & \shortstack{Stage~1\\Compile}
    & FF
    & \shortstack{$A_0$\\Nonzero}
    & \shortstack{$A_1$\\Nonzero }\\
    \midrule

    C 
    & 25
    & 2723
    & 31\%
    & 45\%
    & 10
    & 11
    & 52
    & 48
    & 48 \\

    C
    & 26
    & 5639
    & 69\%
    & 0\%
    & 0
    & 0
    & 64
    & 0
    & 0 \\

    C
    & 27
    & 7425
    & 91\%
    & 0\%
    & 0
    & 0
    & 64
    & 0
    & 0 \\
    \midrule

    D
    & 25
    & 1313
    & 11\%
    & 80\%
    & 38
    & 39
    & 23
    & 64
    & 57 \\

    D
    & 26
    & 7041
    & 86\%
    & 0\%
    & 0
    & 0
    & 64
    & 0
    & 0 \\

    D
    & 29--31
    & 8192
    & 100\%
    & 0\%
    & 0
    & 0
    & 64
    & 0
    & 0 \\
    \bottomrule
    \end{tabular}
\end{table*}

From step 26 onward, both C and D enter an all-zero execution regime.
All 64 trajectories in the observed training iteration fail at both stages, so the
fixed-zero execution score gives identical zero returns within the group.
The resulting group-relative advantages, policy loss, and gradient norm are
therefore zero, while Stage~1 generations become substantially longer and
normal document termination disappears.

At the corresponding steps, A and B remain executable and retain nonzero
training signals on the same sample groups.
C and D use different Stage~0 credits but share the absolute Stage~1 definition,
making it the common credit choice associated with the observed early failure
regime.

\subsection{Summary}
\label{app:credit_summary}

The ablations and fixed-trajectory audit show that the two credit choices affect
different parts of learning.
Using downstream Stage~0 credit changes which initial programs are favored
according to the quality reached after the corrective action.
Using improvement-relative Stage~1 credit produces substantially different
corrective preferences from absolute final-score credit, changing advantage
signs and rankings for a large fraction of the same sampled trajectories.

These differences also appear in training behavior.
Variant A combines downstream Stage~0 credit with improvement-relative Stage~1
credit and completes the full training run, while the alternative variants
exhibit distinct correction behavior or optimization failures.

\section{State-Dependent Correction and Post-Execution Behavior}
\label{app:workflow_behavior}

This section expands the state-dependent correction analysis in
Section~\ref{sec:state_dependent}.
We first report the complete statistics behind
Figure~\ref{fig:state_dependent}, then examine the same behavior across nearby
checkpoints.
We finally analyze revision and repair routing, execution transitions, and the
optional selection rule applied when both outputs are available.

\subsection{Full State-Dependent Correction Analysis}
\label{app:state_dependence}

Figure~\ref{fig:state_dependent} analyzes the Dev-selected checkpoint at
step 100.
As in the main text, we restrict this analysis to paired-success examples for
which both Stage~0 and Stage~1 produce valid renders.
We partition examples according to their Stage~0 RSim and define
\[
\Delta M = M(I_1,I^\star)-M(I_0,I^\star).
\]
We separately report
\[
P(S_1>S_0),
\]
which denotes the fraction of examples with a strictly higher Stage~1 RSim.
This quantity uses the threshold $\Delta\mathrm{RSim}>0$, whereas the
correction-outcome categories used in Section~\ref{sec:workflow_correction}
use a $\pm0.005$ neutral region.

\begin{table*}[h]
\caption{
Complete state-dependent correction statistics for our method at checkpoint 100
on Dev-150.
The analysis is restricted to paired-success examples.
$\Delta$ values denote Stage~1 minus Stage~0.
The $95\%$ intervals are bootstrap intervals over examples within each bucket.
Improve, neutral, and degrade use the $\pm0.005$ RSim threshold from the main
learning-dynamics analysis, whereas $P(S_1>S_0)$ uses the strict
$\Delta\mathrm{RSim}>0$ criterion.
}
\label{tab:state_dependence_full}
\centering
\small
\setlength{\tabcolsep}{6pt}
\begin{tabular}{lcccccccc}
\toprule
Stage~0 RSim
& $n$
& $\Delta$RSim
& $95\%$ CI
& $P(S_1>S_0)$
& Improve
& Neutral
& Degrade \\
\midrule
$[0.0,0.4)$
& 17
& $+0.0260$
& $[+0.0103,+0.0442]$
& 76.5\%
& 58.8\%
& 29.4\%
& 11.8\% \\

$[0.4,0.5)$
& 29
& $+0.0029$
& $[-0.0103,+0.0168]$
& 55.2\%
& 41.4\%
& 20.7\%
& 37.9\% \\

$[0.5,0.6)$
& 39
& $+0.0021$
& $[-0.0087,+0.0128]$
& 53.8\%
& 48.7\%
& 7.7\%
& 43.6\% \\

$[0.6,0.7)$
& 26
& $-0.0026$
& $[-0.0287,+0.0194]$
& 53.8\%
& 53.8\%
& 15.4\%
& 30.8\% \\

$[0.7,1.0]$
& 30
& $-0.0235$
& $[-0.0593,+0.0026]$
& 23.3\%
& 20.0\%
& 33.3\%
& 46.7\% \\
\bottomrule
\end{tabular}
\end{table*}

The full statistics reinforce the pattern shown in the main figure.
The weakest intermediate states receive the largest average correction:
for Stage~0 RSim below $0.4$, Stage~1 improves RSim by $0.0260$ on average and
achieves a higher RSim in $76.5\%$ of paired examples.
The average gain decreases as Stage~0 quality increases.
For the highest-quality bucket, the mean change becomes $-0.0235$, and only
$23.3\%$ of examples obtain a strictly higher Stage~1 RSim.

The same contrast is visible under metrics that are not used to train the
policy.
Table~\ref{tab:state_dependence_cross_metric} reports the corresponding DINO,
DreamSim, and CLIPImg changes.
All four metrics improve on average in the weakest bucket, whereas all four
decrease in the highest-quality bucket.

\begin{table}[h]
\caption{
Cross-metric Stage~1 changes at checkpoint 100.
RSim is the optimization reward; DINO, DreamSim, and CLIPImg are
evaluation-only metrics.
}
\label{tab:state_dependence_cross_metric}
\centering
\small
\setlength{\tabcolsep}{8pt}
\begin{tabular}{lrrrr}
\toprule
Stage~0 RSim
& $\Delta$RSim
& $\Delta$DINO
& $\Delta$DreamSim
& $\Delta$CLIPImg \\
\midrule
$[0.0,0.4)$
& $+0.0260$
& $+0.0195$
& $+0.0352$
& $+0.0191$ \\

$[0.4,0.5)$
& $+0.0029$
& $+0.0002$
& $-0.0001$
& $+0.0015$ \\

$[0.5,0.6)$
& $+0.0021$
& $-0.0166$
& $-0.0093$
& $-0.0033$ \\

$[0.6,0.7)$
& $-0.0026$
& $-0.0222$
& $-0.0189$
& $+0.0030$ \\

$[0.7,1.0]$
& $-0.0235$
& $-0.0044$
& $-0.0133$
& $-0.0109$ \\
\bottomrule
\end{tabular}
\end{table}

The intermediate buckets are less uniform across metrics, while the endpoint
contrast is consistent:
weak intermediate states leave more room for correction, whereas additional
modification of already strong renders is more likely to degrade the result.

\subsection{State Dependence across Checkpoints}
\label{app:state_checkpoint_robustness}

The main analysis uses step 100 because it is the checkpoint selected on
Dev-150 for final evaluation.
We repeat the same analysis at steps 75 and 125 using identical RSim buckets
and definitions.

Table~\ref{tab:state_checkpoint_robustness} summarizes the weakest and strongest
buckets.
Across all three later checkpoints, weak Stage~0 states receive a larger average
gain and are more likely to obtain a higher Stage~1 RSim than strong Stage~0
states.

\begin{table}[h]
\caption{
State-dependent correction across later checkpoints of our method.
``Weak'' denotes Stage~0 RSim below $0.4$ and ``strong'' denotes Stage~0 RSim
in $[0.7,1.0]$.
All statistics use paired-success examples and the same fixed bucket boundaries.
}
\label{tab:state_checkpoint_robustness}
\centering
\small
\setlength{\tabcolsep}{10pt}
\begin{tabular}{lcccc}
\toprule
Checkpoint
& Weak $\Delta$RSim
& Weak $P(S_1>S_0)$
& Strong $\Delta$RSim
& Strong $P(S_1>S_0)$ \\
\midrule
Step 75
& $+0.0412$
& 58.3\%
& $+0.0045$
& 28.0\% \\

\textbf{Step 100}
& $+0.0260$
& 76.5\%
& $-0.0235$
& 23.3\% \\

Step 125
& $+0.0337$
& 81.3\%
& $-0.0158$
& 28.0\% \\
\bottomrule
\end{tabular}
\end{table}

The exact magnitude changes over training, but the separation between weak and
strong intermediate states is preserved.
At step 75, the strongest bucket still has a small positive average
$\Delta$RSim, but Stage~1 achieves a higher RSim in only $28.0\%$ of those
examples.
At steps 100 and 125, the strongest bucket has a negative average change.

\subsection{Revision and Repair Routing during Training}
\label{app:routing_behavior}

The Stage~1 route is determined independently for every trajectory by its own
Stage~0 execution result.
A valid Stage~0 render enters visual revision; an unsuccessful execution enters
repair.

Across the completed training run, the 1,000 target-image groups contain 8,000
complete trajectories.
Of these, 7,447 trajectories ($93.1\%$) enter visual revision and 553
($6.9\%$) enter repair.
The distribution across groups is shown in
Table~\ref{tab:training_route_mix}.

\begin{table}[h]
    \caption{
        Routing composition of the 1,000 training groups.
        Each group contains eight independently sampled complete trajectories.
        A mixed group contains at least one revision trajectory and at least one
        repair trajectory.
        }
    \label{tab:training_route_mix}
    \centering
    \small
    \begin{tabular}{lcc}
    \toprule
    Group composition & Groups & Fraction \\
    \midrule
    Revision only & 738 & 73.8\% \\
    Mixed revision / repair & 259 & 25.9\% \\
    Repair only & 3 & 0.3\% \\
    \midrule
    All groups & 1,000 & 100\% \\
    \bottomrule
    \end{tabular}
\end{table}

The 259 mixed groups also show that a training group contains independently
sampled Stage~0 actions rather than one shared Stage~0 state followed by
multiple corrective candidates.
Each Stage~1 action operates on the execution state produced by its own Stage~0
program.

\subsection{Execution Transitions and Repair}
\label{app:repair_analysis}

We next examine how the corrective action changes execution status.
We denote the four possible transitions by
$\mathrm{SS}$, $\mathrm{SF}$, $\mathrm{FS}$, and $\mathrm{FF}$.
The $\mathrm{FS}$ transition corresponds to execution repair, whereas
$\mathrm{SF}$ denotes an initially valid program that becomes invalid after
correction.

\begin{table*}[h]
\caption{
Execution transitions for the initializer and the Dev-selected checkpoint of our method.
``Repair success'' is $P(S_1\text{ succeeds}\mid S_0\text{ fails})$.
}
\label{tab:execution_transitions}
\centering
\small
\setlength{\tabcolsep}{10pt}
\begin{tabular}{llcccccc}
\toprule
Split
& Policy
& SS
& SF
& FS
& FF
& Stage~0 failures
& Repair success \\
\midrule

Dev-150
& Initializer
& 123 & 0 & 1 & 26
& 27
& 3.7\% \\

Dev-150
& Step 100
& 141 & 4 & 1 & 4
& 5
& 20.0\% \\
\midrule

Final-945
& Initializer
& 780 & 5 & 6 & 154
& 160
& 3.8\% \\

Final-945
& Step 100
& 863 & 13 & 14 & 55
& 69
& 20.3\% \\
\bottomrule
\end{tabular}
\end{table*}

Training substantially improves Stage~0 executability, reducing Final-945
Stage~0 failures from 160 for the initializer to 69 at checkpoint 100.
Among the remaining failed states, Stage~1 repairs 14 of 69 programs, compared
with 6 of 160 for the initializer.
The improvement is therefore already visible before correction is invoked,
while the second action adds repair capability for the failed states that
remain.

The opposite $\mathrm{SF}$ transition also remains possible.
On Final-945, 13 initially successful Stage~0 programs fail after Stage~1.
This motivates retaining the initial output when the corrective action makes
execution worse.

\subsection{Execution-Aware Selection}
\label{app:selection_behavior}

For correction-enabled inference, both Stage~0 and Stage~1 are available.
Because the effect of Stage~1 depends on the intermediate state, we additionally
evaluate the selection rule defined in Appendix~\ref{app:inference_protocol}.
The rule chooses only between the two programs already produced and requires no
additional model generation.

Table~\ref{tab:selection_dev_final} compares Stage~0, raw Stage~1, and the
selected output at the fixed checkpoint 100.

\begin{table*}[h]
\caption{
Stage~0, raw Stage~1, and execution-aware selection for our method at
checkpoint 100.
Visual metrics use the fixed-zero convention.
}
\label{tab:selection_dev_final}
\centering
\small
\setlength{\tabcolsep}{10pt}
\begin{tabular}{llccccc}
\toprule
Split
& Output
& Render
& RSim
& DINO
& DreamSim
& CLIPImg \\
\midrule

Dev-150
& Stage~0
& 96.67\%
& 0.5479
& 0.8225
& 0.7554
& 0.8668 \\

& Stage~1
& 94.67\%
& 0.5358
& 0.7994
& 0.7348
& 0.8485 \\

& Selected
& \textbf{97.33\%}
& \textbf{0.5646}
& \textbf{0.8326}
& \textbf{0.7677}
& \textbf{0.8774} \\
\midrule

Final-945
& Stage~0
& 92.70\%
& 0.5382
& 0.8014
& 0.7330
& 0.8372 \\

& Stage~1
& 92.80\%
& 0.5446
& 0.8046
& 0.7401
& 0.8409 \\

& Selected
& \textbf{94.18\%}
& \textbf{0.5649}
& \textbf{0.8221}
& \textbf{0.7593}
& \textbf{0.8583} \\
\bottomrule
\end{tabular}
\end{table*}

On Dev-150, Stage~0 is already strong at checkpoint 100, while raw Stage~1 is
lower on the aggregate metrics.
The state-dependent analysis nevertheless shows that the corrective action
remains useful for weaker intermediate states.
Selection retains those gains while keeping Stage~0 when it is preferable,
raising selected RSim-FZ to $0.5646$.

On Final-945, raw Stage~1 is slightly higher than Stage~0 under all four visual
metrics, and selection provides a further improvement:
render success rises from $92.80\%$ for raw Stage~1 to $94.18\%$, while
RSim-FZ increases from $0.5446$ to $0.5649$.
As reported in Appendix~\ref{app:bootstrap}, the paired bootstrap interval for
each Stage~1-to-Selected improvement excludes zero on the fixed Final-945
sample set.

Stage~0 remains the single-call output of the trained policy.
The corrective action and selection provide an optional additional inference
mode when execution and a second model call are available.

\subsection{Summary}
\label{app:workflow_behavior_summary}

The extended analysis supports three observations.
First, correction is state dependent: weak intermediate renders receive larger
gains, while strong renders are more susceptible to degradation after further
modification.
Second, the same separation appears across nearby checkpoints rather than only
at the Dev-selected checkpoint.
Third, training improves the initial action while also producing visual
revision and execution-repair behavior in the corrective action; this second
action can be used optionally at inference time.

\end{document}